\documentclass{article}
\usepackage{iclr2027_conference,times}

\newif\ifpreprint
\iclrfinalcopy
\preprinttrue

\newif\ifreview \reviewtrue

\usepackage[T1]{fontenc}
\usepackage[scale=.98]{zlmtt}
\usepackage{microtype}   
\usepackage{mymath}      

\usepackage{calc}
\usepackage{titlesec}
\usepackage{enumitem}
\usepackage{xcolor}
\usepackage[normalem]{ulem}
\usepackage{xspace}
\usepackage{booktabs}
\usepackage{multirow}      
\usepackage{pifont}        
\usepackage{caption}
\usepackage{subcaption}    
\usepackage{algorithm}
\usepackage{algpseudocode}
\usepackage{tikz}
\usetikzlibrary{arrows.meta,positioning,fit,backgrounds,calc}
\usepackage{wrapfig}
\usepackage{mwe}           
\usepackage{listings}      
\usepackage{url}
\usepackage{hyperref}
\usepackage{cleveref}

\titleformat{\paragraph}[runin]{\normalsize\bfseries}{}{0pt}{}[.]
\titlespacing*{\paragraph}{0pt}{1.5ex}{1ex}

\definecolor{cblue}{rgb}{0.169,0.351,0.850}
\hypersetup{colorlinks=true, linkcolor=cblue, citecolor=cblue, urlcolor=cblue}

\newlength{\seclabelindent}
\AtBeginDocument{\settowidth{\seclabelindent}{\large\sc 1\quad}}
\newlength{\oldparindent}
\setlist[itemize]{
  labelsep=1ex,
  leftmargin=\seclabelindent,
  listparindent=\parindent,
  parsep=\parskip,
  topsep=0pt,
  itemsep=0pt,
  partopsep=1ex,
  label=\raisebox{0.25ex}{\scalebox{0.75}{\textcolor{black}{$\bullet$}}}
}
\newcommand\desclabel[1]{#1}
\setlist[description]{
  labelsep=\widthof{~},
  leftmargin=\oldparindent,
  listparindent=\parindent,
  parsep=\parskip,
  topsep=0pt,
  itemsep=0pt,
  partopsep=1ex,
  format=\normalfont\bfseries\desclabel,
}
\setlist[enumerate]{
  labelsep=0.5ex,
  leftmargin=\oldparindent,
  listparindent=\parindent,
  parsep=\parskip,
  topsep=0pt,
  itemsep=0pt,
  partopsep=1ex,
}
\setlist[enumerate,1]{label=\textcolor{black}{\arabic*.}}
\setlist[enumerate,2]{label=\textcolor{black}{\alph*)}}
\setlist[enumerate,3]{label=\textcolor{black}{\roman*.}}

\newcommand{\ours}{\textsc{BlackSwan}\xspace}
\newcommand{\react}{\textsc{ReAct}\xspace}
\newcommand{\accord}{\textsc{Accord}\xspace}
\newcommand{\ace}{\textsc{ACE}\xspace}

\newcommand{\cmark}{\ding{51}}   
\newcommand{\xmark}{\ding{55}}   
\newcommand{\dmark}{\smash{\textsuperscript{\dag}}}
\xspaceaddexceptions{\dmark}
\newcommand{\notdefined}{\textcolor{black!45}{\scriptsize N/A}}

\makeatletter
\newcommand{\crefx}[1]{%
  \@ifundefined{r@#1}%
    {\GenericWarning{}{TODO: \noexpand\crefx to undefined label `#1'}%
     \textbf{[\,?\,\texttt{#1}\,]}}%
    {\cref{#1}}%
}
\DeclareRobustCommand\onedot{\futurelet\@let@token\@onedot}
\def\@onedot{\ifx\@let@token.\else.\null\fi\xspace}
\def\eg{\text{e.g}\onedot}

\makeatother

\makeatletter
\newcommand{\newauthor}[2]{%
  \expandafter\newcommand\csname #1\endcsname[1]{%
    \ifreview{\color{#2}##1}\else ##1\fi}%
  \expandafter\newcommand\csname #1ins\endcsname[1]{%
    \ifreview{\color{#2}##1}\else ##1\fi}%
  \expandafter\newcommand\csname #1del\endcsname[1]{%
    \ifreview\textcolor{#2}{\sout{##1}}\fi}%
  \expandafter\newcommand\csname #1cut\endcsname[2][cut]{%
    \ifreview{\color{#2}\textbf{[##1:\,}\itshape ##2\upshape\textbf{\,]}}\fi}%
  \expandafter\newcommand\csname #1sub\endcsname[2]{%
    \ifreview\textcolor{#2}{\sout{##1}\,##2}\else ##2\fi}%
  \expandafter\newcommand\csname #1subkeep\endcsname[3]{%
    \ifreview{\color{#2}\sout{\textcolor{##1}{##2}}}\,\textcolor{#2}{##3}%
    \else ##3\fi}%
}
\makeatother

\newauthor{hcy}{black}
\newauthor{omh}{red}
\definecolor{khyPurple}{RGB}{140, 70, 180}
\newauthor{khy}{khyPurple}

\title{Action Verification through Conjecture and \\ Refutation for Language Model Agents}

\newcommand{\affc}[1]{\textsuperscript{#1},\ \ }

\newlength{\authcol}\newlength{\authtmp}
\ifdim\authtmp>\authcol \fi

\newlength{\affmark}\newlength{\afftmp}
\ifdim\afftmp>\affmark \fi
\newcommand{\affline}[1]{\makebox[\affmark][l]{\textsuperscript{#1}}}

\author{%
\makebox[\authcol][s]{Minhyeon Oh}\affc{1}Chiyeong Heo\affc{1}Geonwoo Noh\affc{1}Hoyoung Kim\affc{2} \\
\bfseries\makebox[\authcol][s]{\textls[20]{Jangwon Son}}\affc{1}Jungseul Ok\textsuperscript{1}\thanks{Corresponding author.} \\[0.7ex]
{\normalfont\affline{1}Pohang University of Science and Technology} \\
{\normalfont\affline{2}National AI Research Lab} \\[0.5ex]
{\normalfont\texttt{\{minhyeon.oh,jungseul.ok\}@postech.ac.kr}}
}

\begin{document}

\makeatletter
\newsavebox{\authprobe}\newlength{\authH}\newlength{\anonH}
\begingroup\let\thanks\@gobble
 \global\sbox\authprobe{\begin{tabular}[t]{l}\bfseries\rule{\z@}{24pt}\@author\end{tabular}}%
\endgroup
\setlength{\authH}{\dimexpr\ht\authprobe+\dp\authprobe\relax}
\sbox\authprobe{\begin{tabular}[t]{l}\bfseries\rule{\z@}{24pt}Anonymous authors\\Paper under double-blind review\end{tabular}}
\setlength{\anonH}{\dimexpr\ht\authprobe+\dp\authprobe\relax}
\makeatother

\maketitle
\ifpreprint\lhead{Preprint}\fi
\ificlrfinal\else
  \vspace*{\dimexpr\authH-\anonH\relax}%
  \footnotetext{\phantom{Corresponding author.}}%
\fi

\begin{abstract}
Language model agents increasingly act on users' behalf by manipulating external environments through tools, often from high-level instructions that leave critical task- and environment-specific context unstated. To prevent resulting misaligned actions, recent work verifies proposed actions before execution, largely by establishing whether the conditions supporting them hold. We argue that this confirmation-oriented view leaves a fundamental blind spot: a wrong action often passes every check that asks whether it is right. This motivates a different question for verification: rather than asking what evidence supports an action, can the agent find evidence that refutes it? We pursue this direction with \ours{}, a refutation-oriented, expect-suspect-test procedure that predicts the intended effect of each state-mutating action, formulates specific and empirically testable hypotheses for how it could be wrong, and tests them using existing observations or additional non-mutating actions, executing the action only if none survives testing. On challenging AppWorld tasks, \ours{} improves success rate over the strongest baseline by 14.9 and 25.2\%p with and without the backbone model's thinking mode, respectively. Even without thinking or in-context demonstrations, it outperforms stronger baselines equipped with them. The gains also transfer beyond AppWorld, with \ours{} improving WebShop performance by 12.7\%p. More broadly, our results suggest that language model agents have an underused capacity to recognize and self-correct misaligned actions, which can be systematically elicited through active conjecture and refutation before execution.
\end{abstract}

\section{Introduction}

A central promise of language model agents is that they can complete delegated tasks autonomously: on behalf of users, they perceive and manipulate user environments through tools~\citep{react,codeact,openhands}. Yet, in practice, users typically specify high-level goals, leaving critical information implicit in their environments~\citep{preferences-implicit,appworld}. Worse, the user-specific environment states and dynamics lie beyond an agent's parametric knowledge~\citep{wma}. Faced with this information gap, agents may take actions that deviate from the user's actual intent~\citep{ambigswe,blind-goal-directedness}. Unlike a flawed chat response, which can simply be ignored, a flawed action changes the environment itself, turning an incorrect inference about unstated context into a potentially hard-to-reverse consequence.

One natural response to this information gap is to equip agents with richer knowledge of their environments~\citep{autoguide,automanual,awm,ace,reasoningbank,gtta}. Yet user environments are open-ended: it is difficult to know in advance which facts will matter to actions that arise later in a task. Such up-front knowledge acquisition therefore faces a basic question of sufficiency---when has the agent learned enough? 

More recent work shifts the focus from understanding the environment in general to verifying individual proposed actions on the fly~\citep{inferact,toolgate,deaction,asking-before-acting,accord}. These methods intervene before execution and establish whether task- and environment-specific conditions for the action hold. This reflects a \emph{confirmation-oriented} view of action readiness: ask what must be established for the proposed action to proceed, then seek evidence for those conditions. But a wrong action can pass every check that asks whether it is right. For example, asked to ``split last night's dinner bill among the attendees'', an agent may confirm the bill amount and the guest list on the event invitation, yet miss that one invitee canceled by message. 

In this work, we argue for a \emph{refutation-oriented} view~\citep{popper1963} of preemptive action verification. The central question is not merely how much evidence verification should gather, but in which direction it should look: should it seek evidence for an action, or evidence \emph{against} it? The confirmation-oriented view takes the former, admitting an action once the conditions required to support it are established. In contrast, the refutation-oriented view takes the latter: it asks how the action \emph{could be wrong}, constructs potential falsifiers, and admits the action only if none survives empirical scrutiny. The action is thus corroborated by withstanding attempted refutation, rather than confirmed by accumulated support.

To operationalize this view, we propose \ours{}\footnote{The name alludes to Popper's classic black-swan example: observing more white swans cannot verify that all swans are white, whereas a single black swan can refute the claim.}, which subjects each state-mutating action to an expect-suspect-test procedure: the action executes only if it withstands the attempt to refute it; otherwise, it is withheld, and the agent drafts a replacement in light of the refuting evidence. Specifically, \textbf{expect} predicts what the proposed action itself would accomplish toward the user's task, yielding an action-local claim that can be challenged without conflating failures of the current write with unfinished parts of the task. Given this claim, \textbf{suspect} raises specific, empirically testable hypotheses about how it could be wrong. We keep this stage separate from testing, following the principle that hypotheses must not depend on the data used to test them~\citep{popper}: if hypotheses were formulated while evidence is gathered, the evidence could shape the very hypotheses it is then used to adjudicate. Finally, \textbf{test} tries to refute the action by seeking the observations these fixed hypotheses predict, drawing on those already in the trajectory or acquiring new ones through non-mutating actions as needed.

Our experiments with \ours{} support the promise of refutation-oriented verification. On the most
challenging tasks of AppWorld~\citep{appworld}, \ours{} outperforms the strongest baseline by
14.9\%p and 25.2\%p in success rate with and without the backbone model's thinking mode, respectively. Even without thinking, \ours{} outperforms every baseline with thinking by 11.7\%p on these tasks, suggesting that the direction of verification matters more than the amount of reasoning. Moreover, even without any in-context demonstration, \ours{} still outperforms every baseline given three by 9.3\%p. Beyond AppWorld, with only its prompts made domain-neutral, \ours{} improves over the same agent without verification on WebShop~\citep{webshop} by 12.7\%p. More broadly, our results suggest that language model agents have an underused capacity to recognize and self-correct misaligned actions. Our approach elicits this capacity through a process of falsification: active conjecture and refutation before execution. Our contributions are:
\begin{itemize}
    \item {We recast preemptive action verification as a question of direction:} whether an agent should seek evidence \emph{for} a proposed action or actively seek evidence \emph{against} it.
    \item {We introduce \ours}, an expect-suspect-test procedure that frames each proposed write as a {conjecture} and executes it only after attempted {refutation}.
    \item {We suggest that the corrective capacity of agents can be systematically elicited at test time}, across models and environments, even under unfamiliar and task-specific conditions.
\end{itemize}

\section{Related Work}

\paragraph{Task-level information acquisition}
One line of work equips agents with missing information before acting. A line of approaches acquire information from the environment: they rely on a dedicated exploration phase~\citep{gtta}, leverage world knowledge models~\citep{qiao2024agent}, learn from synthesized interactions~\citep{learnbyinteract}, or accumulate reusable artifacts such as manuals~\citep{automanual}, guidelines~\citep{autoguide}, workflows~\citep{awm}, and playbooks~\citep{ace}. Another line of approaches resolve missing information through clarification queries~\citep{qian2024tell,askbeforeplan,zhang2025modeling,activedisambig,vijayvargiya2026asking}. Yet both approaches acquire information before a concrete action exists and therefore cannot target the evidence needed to test a particular write.

\paragraph{Trajectory-based verification and correction}
A second line of work intervenes during execution, examining actions against the trajectory. InferAct~\citep{inferact} infers the task the trajectory has pursued and approves actions consistent with the user's instruction; our \textbf{expect} instead states what the pending write should do, as a claim to be falsified rather than a basis for approval. DeAction~\citep{deaction} combines alignment checks with outcome analysis to detect misaligned actions before execution. Other methods revise beliefs during execution, reflecting on the current state relative to the goal~\citep{reflact} or comparing expected outcomes with observations once an action has taken effect~\citep{evu}. Devil's Advocate~\citep{wang2024devil} anticipates failures before execution but checks them only afterward and recovers by backtracking, presuming reversibility. These methods reason over trajectories produced under the very interpretation they aim to correct, and so rarely find evidence against it before execution.

\paragraph{Action-level evidence acquisition}
A third line of work obtains evidence beyond the trajectory before an action executes. Agents may ask the user mid-task~\citep{asking-before-acting,learningtoask,sageagent}, which inherits the limits of asking noted above, or simulate candidate actions with learned world models~\citep{wma}, where evidence remains predicted rather than observed. Other methods probe the environment directly. EnvProbe~\citep{envprobe} allocates a probing budget over uncertain or action-critical beliefs, while AGrail~\citep{agrail} performs action-specific safety checks based on predefined criteria. ACCORD~\citep{accord} intercepts state-changing actions and uses read-only probes to verify their assumptions about constraints, entities, and environment states. \ours{} shares this access to evidence but differs in how the target of verification is formed: it first specifies what the action should achieve and what would falsify that claim, and only then consults the environment. Prior methods gather evidence to support execution or check predefined conditions; \ours{} acquires evidence to challenge the agent's own interpretation before a write occurs.
\begin{figure}[t]
    \centering
    \includegraphics[width=\linewidth]{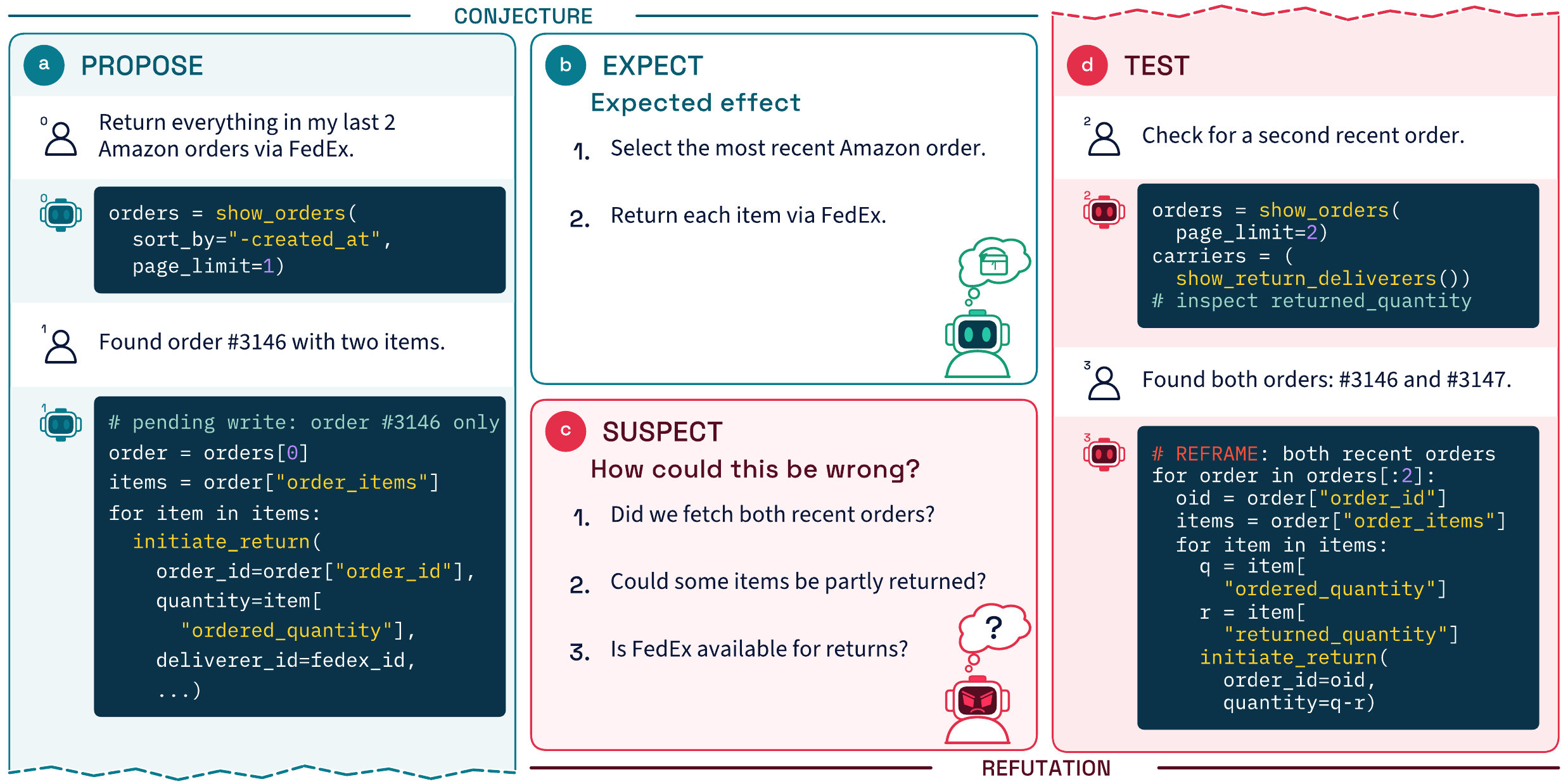}
    \caption{The verification process of \ours.~(a) \textbf{propose} drafts a pending action.~(b) \textbf{expect} states its intended outcome.~(c) \textbf{suspect} lists specific ways the action could be wrong.~(d) \textbf{test} checks these suspicions with read-only probes. If any is confirmed, the action is withheld and reframed using the refuting evidence. The test stage continues directly from the proposal-stage context window.}
    \label{fig:main}
    \vspace{-15pt}
\end{figure}

\providecommand{\method}{\ours}
\providecommand{\accord}{\textsc{Accord}}
\providecommand{\code}[1]{\texttt{#1}}
\providecommand{\doc}{\mathrm{doc}}
\providecommand{\lm}[1]{\mathrm{LM}_{\mathrm{#1}}}

\section{Conjecture and Refutation for Preemptive Verification}
\label{sec:method}

\paragraph{Setup}

Given a natural-language task instruction $u$, an agent interacts with an environment through a set of available APIs. At each step $t \ge 1$, the agent generates an action $a_t \sim \pi(\cdot \mid c_{t-1})$ based on its previous context $c_{t-1}=(u,a_1,o_1,\dots,a_{t-1},o_{t-1})$, with $c_0=(u)$, and receives an observation $o_t$, so that $c_t=c_{t-1}\cat\parens{a_t,o_t}$. We focus on agents that act through executable code~\citep{codeact}: each action $a_t$ is a piece of code executed in a persistent interpreter, and $o_t$ is the output of its execution~(\eg, printed results or errors). A single action may invoke multiple APIs and bind their results to variables, which can be reused both later in the same code and in subsequent actions.

\paragraph{Overview}

We propose \ours{}, a refutation-first approach for preemptive action verification. If an action $a_t$ invokes at least one state-mutating API~(hereafter, a write), its execution is withheld and the action is subjected to our conjecture-and-refutation process; otherwise, it is executed as usual. \ours is inspired by the Popperian view of conjecture and refutation. Rather than asking if $a_t$ appears correct, we ask what observable facts would show $a_t$ to be wrong, and test those possibilities before execution. \Cref{fig:main} illustrates our pipeline: \textbf{expect} makes explicit what $a_t$ claims to accomplish~(\Cref{sec:method:expect}); \textbf{suspect} identifies concrete ways in which that claim could be wrong~(\Cref{sec:method:suspect}); and \textbf{test} determines whether those suspicions are supported by observations of the environment~(\Cref{sec:method:test}).

\subsection{Expect: Making the Action's Intent Explicit}
\label{sec:method:expect}

Given the previous context $c_{t-1}$ and a proposed write $a_t$, the language model predicts what executing $a_t$ would accomplish toward fulfilling the user's task $u$:
\begin{align}
    \kappa_t \sim \pi(\cdot \mid c_{t-1}, a_t) \ .
    \label{eq:expect}
\end{align}
This prediction $\kappa_t$ makes explicit the conjecture behind the proposed action: ``executing $a_t$ should produce $\kappa_t$ in a way that advances the user's task.'' It then serves as the claim that the subsequent \textbf{suspect} stage to challenge.

To make refutation meaningful, we place two constraints on the prediction. First, $\kappa_t$ is generated before any new observation is gathered, so that it cannot be shaped to fit the very evidence used to test it. Second, $\kappa_t$ is kept action-local: it is intended to capture effects attributable to the proposed action itself, rather than broader outcomes of the overall task execution.

\subsection{Suspect: Constructing Potential Falsifiers}
\label{sec:method:suspect}

Given the previous context $c_{t-1}$, the proposed write $a_t$, and its prediction $\kappa_t$, the language model identifies concrete ways in which the conjecture could fail, producing a set of suspicions
\begin{equation}
    S_t = \braces{s_1, \ldots, s_k} \sim \pi(\cdot \mid c_{t-1}, a_t, \kappa_t) \ ,
    \label{eq:suspect}
\end{equation}
where each $s_i$ describes a condition under which executing $a_t$ could lead to an effect inconsistent with $\kappa_t$ or with the user's task $u$. These suspicions serve as potential falsifiers, rather than conclusions that the action is wrong. The subsequent \textbf{test} stage determines from observations whether any of them actually holds.

\begin{wrapfigure}{r}{0.38\textwidth}
    \vspace{-\intextsep}
    \centering
    \includegraphics[width=\linewidth]{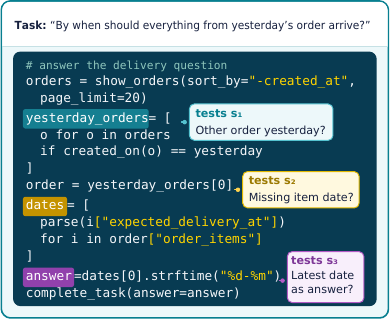}
    \captionsetup{skip=1.2ex}
    \caption{Code-grounded suspicions identify testable assumptions that may change the final answer.}
    \label{fig:provenance}
\end{wrapfigure}
\setlength{\columnsep}{3ex}
\paragraph{Consequential and testable suspicions}
Our goal is not to surface every possible error in the prior context $c_{t-1}$ or the proposed write $a_t$, but only those that~\textbf{(1)} could affect the pending write and~\textbf{(2)} can be empirically tested against observations. A suspicion may concern any part of $c_{t-1}$ or $a_t$, but its effect must \emph{reach} the write action \emph{through some concrete code element} $z_i$ of $a_t$, such as local variables and API parameters. To operationalize these requirements, we require the model to emit each suspicion $s_i$ together with the corresponding element as a pair $(s_i, z_i)$, encouraging it to reason about the provenance of $z_i$ and the ambient state surrounding the associated API call~(\Cref{fig:provenance}). As a result, the suspicions focus on consequential failures rather than on the potentially unbounded set of concerns that never affect the proposed action, while remaining amenable to empirical testing.
\setlength{\columnsep}{10pt}

\paragraph{Preregistered suspicions}
We deliberately separate \textbf{suspect} from \textbf{test}, rather than letting the model raise and check concerns in a single interleaved loop. When the two are combined, the model tends to raise only concerns it expects to resolve easily~\citep{popper}, so the hypotheses end up shaped by the very observations meant to test them. Instead, $S_t$ is generated before any probing, from $c_{t-1}$, $a_t$, and $\kappa_t$ alone as in \eqref{eq:suspect}: at this stage the model reasons only over the proposed code and what has already been observed, without issuing any new action. Much like preregistration in empirical science, where hypotheses are committed before data collection, this fixes what would count as a refutation before any evidence is gathered. It also holds \textbf{test} accountable: since \textbf{test} must reach a verdict on every suspicion in $S_t$, each registered suspicion is explicitly confirmed, dismissed, or left unresolved, and none can be silently dropped or reframed once probing begins.

\subsection{Test: Trying to Refute the Conjecture}
\label{sec:method:test}

Given the previous context $c_{t-1}$, the proposed write $a_t$, and the potential falsifiers $S_t$, the language model attempts to resolve each suspicion $s_i \in S_t$ using observations from the environment. Rather than judging a suspicion from the proposed code and reasoning alone, the model must resolve each $s_i$ from observable evidence: it can revisit observations already available in $c_{t-1}$ or acquire new ones as needed by issuing non-mutating actions.

\paragraph{Probing and verdict}

The pending write $a_t$ is appended to $c_{t-1}$ as an unexecuted action, after which the agent receives an observation $o_t$ that presents $S_t$ as suspicions to resolve, yielding $c_t=c_{t-1}\cat\parens{a_t,o_t}$. Concretely, $o_t$ is a user-role message of the form ``I'm suspicious of executing this action for the following reasons: $S_t$. Please verify whether each of them is the case.'' The agent then continues from $c_t$. At each probing step $k \geq 1$, it either issues a non-mutating action $a_{t+k} \sim \pi(\cdot \mid c_{t+k-1})$, whose observation $o_{t+k}$ extends the context as $c_{t+k} = c_{t+k-1} \cat \parens{a_{t+k}, o_{t+k}}$, or emits a verdict $v_t\colon S_t\to\braces{\true,\false,\none}$ that resolves each suspicion as follows:
\begin{align}
v_t(s_i)=
\begin{cases}
\true  & \text{if observations confirm } s_i \  ,\\
\false & \text{if observations dismiss } s_i \  ,\\
\none  & \text{otherwise} \ ,
\end{cases}
\end{align}
where $\none$ covers suspicions that probing can neither confirm nor dismiss, typically because they depend on the outcome of executing $a_t$ itself~(\eg, a transient network failure during an API call). Testing ends when the agent emits $v_t$ at some step $K$. Unlike the probing actions, $v_t$ is not appended, leaving the context at $c_{t+K-1}$.

\paragraph{Acting on the verdict}

The verdict determines whether the pending write is released as follows:
\begin{subequations}\label{eq:decision}
\begin{alignat}{2}
    &\forall s_i \in S_t\colon\; v_t(s_i)=\false \lor v_t(s_i)=\none
    &\quad& \Rightarrow\quad \text{execute } a_t,
    \label{eq:execute}\\
    &\exists s_i \in S_t\colon\; v_t(s_i)=\true
    && \Rightarrow\quad \text{withhold } a_t \text{ and draft } a_t'.
    \label{eq:withhold}
\end{alignat}
\end{subequations}
In \eqref{eq:execute}, the proposed action $a_t$ has survived every refutation attempt; suspicions labeled $\none$ do not block execution because they cannot be resolved without observing the effect of $a_t$. Thus, $a_t$ is reissued as the next action, $a_{t+K}=a_t$, this time executed. In \eqref{eq:withhold}, a single confirmed suspicion suffices to withhold $a_t$, as the action is no longer consistent with the conjecture it was intended to fulfill. The agent drafts a replacement $a_t'$ from $c_{t+K-1}$ with $v_t$ provided as additional context~(\eg., as an assistant prefill). If $a_t'$ is another write action, it undergoes the same expect-suspect-test procedure. In either case, the context then grows as before, $c_{t+K}=c_{t+K-1}\cat\parens{a_{t+K},o_{t+K}}$.

\section{Experiments}
\label{sec:experiments}
Unless otherwise stated, all methods use the official FP8 checkpoint of Qwen3.5-27B~\citep{qwen3.5} as the underlying model, following \accord{}~\citep{accord}.

\paragraph{Benchmarks}
Our primary benchmark is AppWorld~\citep{appworld}, where an agent with access to 457 APIs across 9 simulated applications completes user requests by discovering and invoking the ones it needs. AppWorld captures key challenges of assistant deployment: agents must discover persistent user data, avoid unintended state changes, and account for irreversible writes. Tasks are grouped into scenarios of three similar variants that differ in user and environment states. On the normal~(168 tasks) and challenge~(417 tasks) test splits, we report the two official metrics: \emph{task goal completion}~(TGC), the fraction of tasks that pass all of their unit tests~(8 on average, up to 22), and \emph{scenario goal completion}~(SGC), the fraction of scenarios in which all three tasks pass. We additionally evaluate on WebShop~\citep{webshop} and ALFWorld~\citep{alfworld} to show that \ours{} generalizes to environments with different action spaces.

\paragraph{Baselines}
\react{}~\citep{react} interleaves reasoning with code actions~\citep{codeact} and executes every action immediately, serving as the unguarded reference on which \ours{} and the other baselines are built. \ace{}~\citep{ace} enriches the agent's knowledge of its environment with a playbook of strategies and pitfalls distilled from training tasks.\footnote{We run its released implementation (\url{https://github.com/ace-agent/ace}).} Comparing \ours{} against it tests whether knowledge supplied in advance suffices, or whether erroneous mutations also stem from task-specific facts that it cannot anticipate. \accord{}~\citep{accord} admits each write once revisited or newly probed observations support its required conditions.\footnote{As no implementation of \accord{} was publicly available as of September 25, 2026, we reimplemented it from the paper, resolving unspecified design choices as consistently with its description as possible~(\Cref{app:accord}).} As an instance of the confirmation-oriented view, it isolates the direction in which verification looks: \accord{} seeks evidence for the write, whereas \ours{} seeks evidence against it.

\subsection{Main Results}
\label{sec:exp:results}
\begin{figure}[t]
\begin{minipage}[t]{0.281\linewidth}\vspace{0pt}
  \centering
  \includegraphics[width=\linewidth]{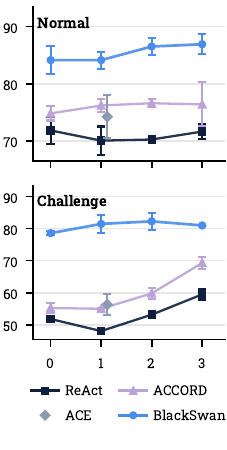}
\end{minipage}\hfill
\begin{minipage}[t]{0.695\linewidth}\vspace{0pt}
  \captionsetup{type=table}
  \caption{Results on the normal and challenge splits of AppWorld, with all methods using a single in-context task examples. Mean $\pm$ standard deviation over three runs. We report results with the backbone model's thinking on~(\cmark) and off~(\xmark). $^{\dagger}$Reported in
  the original paper; its number of in-context demonstrations is unspecified.}
  \label{tab:onedemo}
  \centering\small\setlength{\tabcolsep}{3pt}
  \begin{tabular*}{\linewidth}{@{\extracolsep{\fill}}clllll@{}}
\toprule
 & & \multicolumn{2}{l}{Test (normal)} & \multicolumn{2}{l}{Test (challenge)} \\
\cmidrule(l{1pt}r{1pt}){3-4}\cmidrule(l{1pt}r{1pt}){5-6}
 & Method & TGC & SGC & TGC & SGC \\
\midrule
\multirow{5}{*}{\cmark{}} & \react{} & 72.4{\scriptsize$\,\pm$1.9} & 56.5{\scriptsize$\,\pm$3.7} & 58.4{\scriptsize$\,\pm$0.8} & 43.2{\scriptsize$\,\pm$2.5} \\
    & \ace{} & 80.2{\scriptsize$\,\pm$3.8} & \underline{68.2}{\scriptsize$\,\pm$4.8} & 65.5{\scriptsize$\,\pm$4.2} & 49.6{\scriptsize$\,\pm$4.3} \\
    & \accord{}\dmark{} & \underline{81.5}\phantom{{\scriptsize$\,\pm$0.0}} & 67.9\phantom{{\scriptsize$\,\pm$0.0}} & \underline{69.8}\phantom{{\scriptsize$\,\pm$0.0}} & \underline{53.2}\phantom{{\scriptsize$\,\pm$0.0}} \\
    & \accord{} & 79.4{\scriptsize$\,\pm$1.8} & 65.5{\scriptsize$\,\pm$5.7} & 63.5{\scriptsize$\,\pm$1.7} & 48.4{\scriptsize$\,\pm$1.5} \\
\cmidrule{2-6}
    & \ours{} & \textbf{88.3}{\scriptsize$\,\pm$3.0} & \textbf{76.8}{\scriptsize$\,\pm$5.4} & \textbf{84.7}{\scriptsize$\,\pm$1.3} & \textbf{72.4}{\scriptsize$\,\pm$2.2} \\
\midrule
\multirow{4}{*}{\xmark{}} & \react{} & 70.0{\scriptsize$\,\pm$2.5} & 48.2{\scriptsize$\,\pm$5.4} & 48.0{\scriptsize$\,\pm$0.8} & 33.6{\scriptsize$\,\pm$1.1} \\
    & \ace{} & 74.2{\scriptsize$\,\pm$3.8} & 57.7{\scriptsize$\,\pm$5.7} & \underline{56.3}{\scriptsize$\,\pm$3.1} & 41.8{\scriptsize$\,\pm$3.4} \\
    & \accord{} & \underline{76.2}{\scriptsize$\,\pm$1.2} & \underline{59.5}{\scriptsize$\,\pm$2.7} & 55.0{\scriptsize$\,\pm$0.4} & \underline{44.8}{\scriptsize$\,\pm$2.3} \\
\cmidrule{2-6}
    & \ours{} & \textbf{84.1}{\scriptsize$\,\pm$1.5} & \textbf{69.6}{\scriptsize$\,\pm$1.8} & \textbf{81.5}{\scriptsize$\,\pm$2.9} & \textbf{65.0}{\scriptsize$\,\pm$4.1} \\
\bottomrule
\end{tabular*}

\end{minipage}%
\captionsetup{skip=-5pt}
\caption{TGC~(vertical, \%) versus the number of in-context demonstrations~(horizontal, 0 to 3), with the backbone model's thinking mode off, on the normal~(top) and challenge~(bottom) splits. Error bars show the standard deviation over three runs.}
\label{fig:shots-nothink}
\end{figure}

\paragraph{Overall performance}
\Cref{tab:onedemo} shows that \ours{} achieves the highest performance on AppWorld, regardless of whether the underlying model's thinking mode is on or off. With thinking on, it improves over the strongest baseline in each column~(underlined) by 6.8 TGC and 8.6 SGC points on the normal split, and by 14.9 and 19.2 points on the challenge split. Without thinking, the corresponding margins are 7.9 and 10.1 points on the normal split, and 25.2 and 20.2 points on the challenge split. As \ace{} and \accord{} represent the two prevailing responses to the information gap, namely knowledge acquired in advance and confirmation-oriented verification, these margins suggest that subjecting each proposed action to conjecture and refutation resolves errors that both often leave unaddressed.

\setlength{\columnsep}{3ex}
\paragraph{Robustness under distribution shift}
\begin{wrapfigure}{r}{0.36576\linewidth}
\centering
\includegraphics[width=\linewidth]{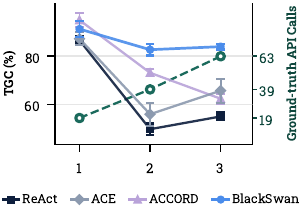}
\captionsetup{skip=2ex}
\caption{AppWorld results by difficulty, from 1~(easiest) to 3~(hardest).}
\label{fig:difficulty}
\end{wrapfigure}%

Moving from the normal to the challenge split in \Cref{tab:onedemo}, the TGC of \ours{} drops by only 3.6 points with thinking on and 2.6 without, whereas even the most robust baseline, \accord\dmark{} with thinking on, drops by 11.7 points. \ace{} and \accord{} improve over \react{} on both splits but drop nearly as much as \react{}. That is, knowledge acquired in advance and confirmation-oriented verification raise performance, but do not reduce the drop on the challenge split. For \ace{}, this is because its playbook is distilled from training tasks, none of which involve the applications required by the challenge split. The gap is also evident in SGC: on the challenge split, \ours{} completes entire scenarios more often than any baseline completes individual tasks, both with thinking on~(72.4 SGC versus at most 69.8 TGC) and off~(65.0 versus 56.3). The gap also widens \emph{within} the challenge split. Across AppWorld's difficulty levels~(\Cref{fig:difficulty}), our margin over \accord{} grows to +21.5 points at the hardest, which requires orchestrating the largest number of API calls.
\setlength{\columnsep}{10pt}

\paragraph{Gains beyond reasoning and demonstrations}
\ours{} outperforms the baselines even when they are given more reasoning or more in-context supervision. In \Cref{tab:onedemo}, \ours{} without thinking outperforms every baseline with thinking, including \accord{}$^{\dagger}$, on all four metrics. \Cref{fig:shots-nothink} varies the number of in-context demonstrations from zero to three, with thinking off. Additional demonstrations benefit the baselines, most notably \accord{} on the challenge split, whose TGC rises from about 55 to 69 points. \ours{}, in contrast, stays within about 3 points across demonstration counts and, even without any demonstration, outperforms every baseline given three. These results suggest that the verification procedure of \ours{} offers benefits that additional reasoning or demonstrations do not readily provide.

\subsection{Transfer to Other Models and Environments}
\label{sec:exp:transfer}
\begin{table}[t]
\centering\small
\caption{\textbf{(a)} More results on AppWorld with Qwen3-4B-Thinking-2507, MiniMax~M2.7, and GLM-5.3-Flash as backbones.~\textbf{(b)} Mean $\pm$ standard deviation of performance on WebShop andALFWorld. The dash~(--) indicates a metric not reported in the paper. N/A indicates a metric not defined by the benchmark. We note that, unlike ours, \accord{} on ALFWorld uses a dedicated procedure.}
\label{tab:generality}
\begin{subtable}[t]{0.60\linewidth}\vspace{0pt}
  \centering\setlength{\tabcolsep}{3pt}
  \caption{AppWorld results with three more backbone models.}\label{tab:backbones}
  \begin{tabular*}{\linewidth}{@{\extracolsep{\fill}}lllllll@{}}
\toprule
& \multicolumn{2}{l}{Qwen3-4B} & \multicolumn{2}{l}{MiniMax} & \multicolumn{2}{l}{GLM} \\
\cmidrule(l{1pt}r{1pt}){2-3}\cmidrule(l{1pt}r{1pt}){4-5}\cmidrule(l{1pt}r{1pt}){6-7}
Method & TGC & SGC & TGC & SGC & TGC & SGC \\
\midrule
\multicolumn{7}{@{}l}{{\itshape Test} (normal)} \\
\react{} & \underline{23.2} & \underline{14.3} & 61.3 & 41.1 & \underline{83.9} & \underline{67.9} \\
\accord{} & 14.9 & 3.6 & \underline{71.4} & \underline{50.0} & 82.1 & 66.1 \\
\cmidrule{1-7}
\ours{} & \textbf{41.7} & \textbf{30.4} & \textbf{77.4} & \textbf{64.3} & \textbf{91.1} & \textbf{82.1} \\
\midrule
\multicolumn{7}{@{}l}{{\itshape Test} (challenge)} \\
\react{} & \underline{12.2} & \underline{4.3} & 53.0 & 35.3 & \underline{77.5} & 55.4 \\
\accord{} & 8.6 & 2.2 & \underline{63.1} & \underline{46.0} & 76.7 & \underline{56.8} \\
\cmidrule{1-7}
\ours{} & \textbf{27.6} & \textbf{13.7} & \textbf{73.9} & \textbf{53.2} & \textbf{83.5} & \textbf{66.9} \\
\bottomrule
\end{tabular*}

\end{subtable}\hfill
\begin{subtable}[t]{0.365\linewidth}\vspace{0pt}
  \centering
  \caption{Two further environments.}\label{tab:otherenvs}
  \begin{tabular*}{\linewidth}{@{\extracolsep{\fill}}lll@{}}
\toprule
& \multicolumn{2}{l}{Performance} \\
\cmidrule(lr){2-3}
Method & Score & SR \\
\midrule
\multicolumn{3}{@{}l}{{\itshape WebShop} (test 500)} \\
\react{} & \underline{55.4}{\scriptsize$\,\pm$1.0} & 38.1{\scriptsize$\,\pm$0.8} \\
Harness-R1 & \makebox[\widthof{55.4}][l]{--}\phantom{{\scriptsize$\,\pm$0.0}} & \underline{42.0}\phantom{{\scriptsize$\,\pm$0.0}} \\
\cmidrule{1-3}
\ours{} & \textbf{72.8}{\scriptsize$\,\pm$0.5} & \textbf{50.8}{\scriptsize$\,\pm$1.2} \\
\midrule
\multicolumn{3}{@{}l}{{\itshape ALFWorld} (unseen)} \\
\react{} & \notdefined & 98.8{\scriptsize$\,\pm$0.9} \\
\accord{} & \notdefined & \underline{99.3}\phantom{{\scriptsize$\,\pm$0.0}} \\
\cmidrule{1-3}
\ours{} & \notdefined & \textbf{99.5}{\scriptsize$\,\pm$0.4} \\
\bottomrule
\end{tabular*}

\end{subtable}
\end{table}

\paragraph{Other backbone models}
\Cref{tab:backbones} repeats the AppWorld comparison with three further models: the dense Qwen3-4B-Thinking-2507~\citep{qwen3technicalreport} and two large mixture-of-experts models, MiniMax~M2.7~\citep{chen2026minimax} and GLM-5.3-Flash~\citep{glm5team2026glm5vibecodingagentic}. With every model, \ours{} outperforms both baselines on all metrics and splits. In contrast, \accord{} improves over \react{} in TGC only with MiniMax~M2.7. With GLM-5.3-Flash and Qwen3-4B, it falls below \react{} on both splits, by up to 8.3 points with Qwen3-4B~(14.9 versus 23.2 on the normal split). These results suggest that the benefit of confirmation-oriented verification depends on the model, whereas our conjecture and refutation improves performance with every model we test.

\paragraph{Other environments}
\Cref{tab:otherenvs} evaluates \ours{} on two benchmarks with different action spaces: WebShop, where the agent navigates an online store through search and click actions~(\eg, \code{click[Buy Now]}), and ALFWorld, where it acts in a household through text commands~(\eg, \code{take apple 1 from countertop 2}). \ours{} runs the same, proposed procedure as on AppWorld, with only its prompts made domain-neutral. On WebShop, it improves the success rate over \react{} by 12.7 points and over Harness-R1~\citep{harnessr1} by 8.8. On ALFWorld, it improves over \react{} by 0.7 points and over \accord{} by 0.2. However, \accord{} here uses a procedure designed specifically for ALFWorld, unlike ours. Note that the applicable baselines differ across benchmarks; for example, \accord{} is defined over write APIs, which WebShop does not expose in the same form. The results in \Cref{tab:otherenvs} therefore test whether \ours{} transfers to other environments, rather than replicating the comparison in \Cref{tab:onedemo}.

\begin{figure}[t]
\centering
\includegraphics{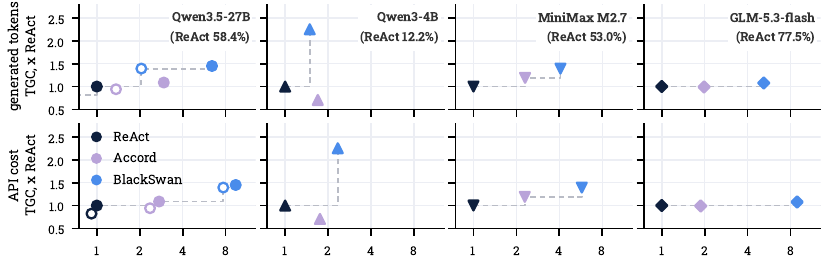}
\caption{Cost~(horizontal) and completion rate~(vertical) on test challenge split of AppWorld, relative to \react{} with thinking on the same backbone, whose completion rate is given in each panel. The top row counts generated tokens only; the bottom row adds input tokens at a common price. Filled markers denote thinking on, and hollow markers thinking off.}
\label{fig:cost-harness}
\end{figure}

\begin{figure}[t]
\centering
\includegraphics[width=\linewidth]{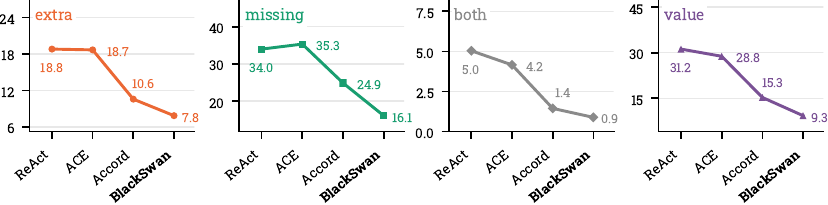}
\caption{Failure types on the challenge split with thinking: collateral changes~(\emph{extra}), required changes left undone~(\emph{missing}), both together~(\emph{both}), and changes to the right items with wrong values~(\emph{value}). All counts are errors per 100 tasks, and an agent may make several errors on a task.}
\label{fig:failurekind}
\end{figure}

\section{Analysis}
\label{sec:analysis}

\label{sec:exp:axes}
\begin{wraptable}{r}{0.365\linewidth}
\vspace{-\intextsep}
\centering\small
\caption{Verification orientation and externalization, varied separately.}
\label{tab:ablation}
\setlength{\tabcolsep}{2pt}
\begin{tabular*}{\linewidth}{@{\extracolsep{\fill}}llll@{}}
\toprule
& \multicolumn{2}{l}{Suspicions} & \\
\cmidrule(l{1pt}r{1pt}){2-3}
Orientation & Explicit & Implicit & $\Delta$ \\
\midrule
Refutation & \textbf{81.6} & 72.7 & $8.9$ \\
Confirmation & 76.0 & 72.6 & $3.4$ \\
\midrule
$\Delta$ & $5.6$ & $0.1$ & \\
\bottomrule
\end{tabular*}

\end{wraptable}
\setlength{\columnsep}{3ex}
\paragraph{Ablation}
\Cref{tab:ablation} varies two design choices of \ours{} separately on test-challenge with thinking
off: whether the verifier asks how the proposed action could be wrong~(refutation) or what it
requires to hold~(confirmation), and whether the potential falsifiers are written out and fixed
before testing~(explicit) or formed by the agent internally as it tests~(implicit). For reference, \accord{}, which conceptually corresponds to confirmation with implicit suspicions, reaches $69.3 \pm 1.8$\% TGC. We find that the two choices are effective only in combination. Refutation improves TGC by 5.6 points when suspicions are explicit but by 0.1 when they are implicit, and fixing suspicions in advance improves TGC by 8.9 points under refutation but by 3.4 under confirmation. This supports the preregistration of suspicions in \Cref{sec:method:suspect}: asking how an action could be wrong improves verification when the resulting suspicions are fixed before testing.
\setlength{\columnsep}{10pt}

\paragraph{Computational cost}
\Cref{fig:cost-harness} compares each agent with \react{} on the same backbone. \accord{} costs
1.7--3.0$\times$ as much as \react{} without consistently improving completion, and falls below
it on Qwen3-4B~($0.71\times$). \ours{} lies on the frontier on every backbone: on Qwen3.5-27B,
Qwen3-4B, and MiniMax M2.7, it completes 1.39--2.25$\times$ as many tasks at 1.5--9.4$\times$
the cost, gaining most on the weakest, Qwen3-4B. On GLM-5.3-flash, where \react{} already
completes 77.5\% of tasks, \ours{} gains less but is the only agent to improve. Both cost
measures rank the agents identically, so the comparison is insensitive to cache reuse.

\paragraph{Failures reduced by verification}
Omission, a required change that the agent never makes, is the most frequent error for every
agent, and it is the error that verifying proposed writes is least able to see, since no write
for it is ever proposed. Both \accord{} and \ours{} reduce every kind of error relative to
\react{}, and \ours{} reduces each more than \accord{}. Per 100 tasks, \ours{} leaves 16.1 required
changes undone, against 24.9 for \accord{}, a reduction of 35\%; it also makes 39\% fewer wrong
values and 26\% fewer collateral changes. In absolute terms, the largest reduction is in omission,
which calls for an explanation.

\setlength{\columnsep}{3ex}
\begin{wrapfigure}{r}{0.3\linewidth}
\vspace{-\intextsep}
\centering
\includegraphics[width=\linewidth]{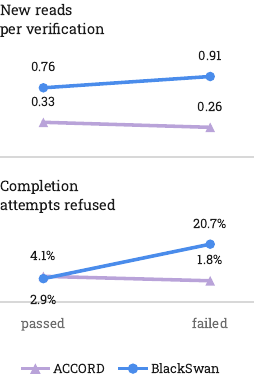}
\label{fig:gate}
\end{wrapfigure}
\paragraph{Burden of proof}
The two verifiers place the burden of proof differently. \accord{} rejects an action only
when an observation confirms a violation and otherwise is told to ``default to
APPROVE''~\citep{accord}, placing the burden on stopping the action; asking what would show
the action wrong places it on executing it. Two consequences follow (right).
First, \ours{} gathers more new evidence: per verification, it makes 0.76 reads the agent had
not already made on completed tasks and 0.91 on failed ones, against 0.33 and 0.26 for
\accord{}. As a result, no verification read reached the faulted application in 53.1\% of
\accord{}'s failures, against 11.3\% for \ours{}. Second, the verdict of \ours{} tracks the
outcome. On the action that ends the task, the last chance to catch an omission, \ours{}
rejects it in 20.7\% of failed tasks and 2.9\% of successful ones, whereas \accord{} rejects
it in 1.8\% and 4.1\%. This reversal is specific to that gate: over all verifications,
\ours{} rejects 54.4\% of actions against 8.4\% for \accord{}, and both reject more often on
failed trials.\\\mbox{}
\vspace{-\baselineskip}
\setlength{\columnsep}{10pt}

\label{sec:exp:cost}

\section{Conclusion}

We introduced a refutation-oriented approach to preemptive action verification, where agents actively seek evidence against proposed state-mutating actions before execution. \ours{} operationalizes this idea with an expect-suspect-test procedure that fixes testable suspicions before gathering evidence. On AppWorld, \ours{} achieves 84.7 TGC and 72.4 SGC on the challenge split, improving over the strongest baseline by 14.9 and 19.2 points, with even larger TGC gains when thinking is disabled; it also remains stronger without demonstrations than baselines given three. The gains are consistent across four backbones and transfer to WebShop and ALFWorld, while ablations show that generating suspicions helps substantially only when they are fixed before testing. Overall, the results suggest that agents have an underused capacity to recognize and correct misaligned actions at test time through active conjecture and refutation before execution.

\paragraph{Limitations and future work}
Refutation cannot resolve genuine ambiguity: when user instructions admit multiple plausible interpretations, it may repeatedly surface the same unresolved issue across successive proposals, increasing verification cost. A natural extension is to treat such recurring suspicions as a signal of ambiguity and escalate them to the user for clarification.

\newpage
\subsection*{AI use statement}
\hcy{In this work, we used generative AI tools (Claude Code) to implement methods, to design and give feedback on
experiments, to propose and refine hypotheses, to support qualitative analysis of agent trajectories, and to
interpret results. We have not used generative AI tools to generate synthetic data sets, to develop the conceptual
framework, to assist with translation, or to clean or reformat data sets, and formulating mathematical claims or
proofs is not applicable to this work. We have reviewed all AI-assisted work. We take responsibility for the final
content of this work, including text, claims or artifacts produced with the aid of generative AI.}

\subsection*{Ethics statement}
\hcy{All experiments run in the simulated environments of AppWorld~\citep{appworld}, WebShop~\citep{webshop}, and ALFWorld~\citep{alfworld}. They involve no human subjects and no real personal data, as the users in AppWorld are fictitious. Pre-execution verification reduces but does not eliminate erroneous agent
actions, so agents acting on real accounts still require further safeguards. Models accessed through a commercial
API received only benchmark content.}

\subsection*{Reproducibility statement}
\hcy{Implementation details are given in the appendix: our agent (\Cref{app:impl-ours}) and its prompts
(\Cref{app:prompts-ours}), the reproduction of \accord{} and its prompts
(\Cref{app:reproduce-accord,app:prompts-accord}), and cost measurement (\Cref{app:cost}).}

\newpage
\appendix
\raggedbottom

\section{Implementation details of \ours{}}
\label{app:impl-ours}

\algnewcommand{\OursIf}[2]{\State \algorithmicif\ #1\ \algorithmicthen\ #2}

\hcy{\Cref{alg:ours} gives the loop of \ours{} on one task, and \Cref{tab:ours-limits} its limits. \ours{} is a
code agent in the style of \react{}: each step drafts one Python cell $c$ and runs it. $h$ is the history the
model sees, and $h \oplus (\cdot)$ appends turns to it. Each call marked with a prompt reads $h$ but adds nothing
to it; only the appends in \Cref{alg:ours} do.}

\begin{algorithm}[H]
\caption{\ours{} on one task.}
\label{alg:ours}
\begin{algorithmic}[1]
\Require instruction $x$, model $M$
\State $h \gets$ the \react{} prompt with $x$; \ $T \gets \emptyset$ \Comment{word--endpoint tables}
\Repeat
  \For{$a = 0, \ldots, 3$} \Comment{at most 3 redrafts}
    \State $c \gets \textsc{Draft}(M, h)$ \Comment{Prompt~\ref{lst:ours-draft}; form rules}
    \OursIf{$c$ only reads}{\textbf{break}}
    \State $e \gets \textsc{Expect}(M, h, c)$ \Comment{Prompt~\ref{lst:ours-expect}}
    \State $S \gets \textsc{Suspect}(M, h, c, e, \textsc{Anatomy}(c), T)$
      \Comment{Prompts~\ref{lst:ours-suspect}, \ref{lst:ours-suspect-finish}}
    \OursIf{$S = \emptyset$}{\textbf{break}}
    \State $(P, V) \gets \textsc{Test}(M, h, c, S)$ \Comment{Prompt~\ref{lst:ours-test}}
    \State $h \gets h \oplus (c, S, P, V)$ \Comment{probes $P$, verdicts $V$}
    \OursIf{no verdict in $V$ is \emph{confirmed}}{\textbf{break}}
  \EndFor
  \State $o \gets$ run $c$; \ $h \gets h \oplus (c, o)$ \Comment{the last draft runs even if blocked}
  \OursIf{$c$ lists a new app $u$'s endpoints}{$T \gets T \cup \textsc{Ground}(M, h, u, x)$}
    \Comment{Prompt~\ref{lst:ours-ground}}
\Until{$c$ calls \code{complete\_task}}
\end{algorithmic}
\end{algorithm}

\hcy{\paragraph{Draft}
Three rules check the form of a drafted cell. A cell that calls an endpoint whose documentation the agent has not
read is replaced by that lookup; a cell that reads an endpoint's documentation before the app's endpoint list is
replaced by the list; and a cell that calls \code{complete\_task} together with other code is drafted again. A
cell \emph{only reads} when every API call in it is a GET request, a documentation lookup, or \code{login}; a cell
with no API call is verified.

\paragraph{Verification}
\textsc{Anatomy} is a static analysis of the cell: its variable bindings and API calls in evaluation order, each
call with every parameter, including those left at their documented defaults. On a cell that calls
\code{complete\_task}, \textsc{Suspect} asks instead what part of the task is not done yet
(Prompt~\ref{lst:ours-suspect-finish}), at most twice per task. \textsc{Test} judges all suspicions together. At
each turn the model either writes a probe, which the harness runs and answers with its output, or ends with a
verdict for every suspicion: \emph{confirmed}, \emph{dismissed}, or \emph{deferred}. A probe may only read. A probe
that makes the cell's own write defers every verdict, since only running the cell can settle it; a probe that makes
any other write is refused, and the model is asked again. Only a confirmed suspicion blocks the cell. The verdicts
then end $h$, so the next draft continues from them.

\paragraph{Context}
Before each model call, while $h$ exceeds the context budget, the harness shortens its cell outputs one step at a
time, oldest first; every value it clips stays retrievable from the environment.}

\begin{table}[H]
\centering
\small
\caption{Limits of \ours{} in the runs of \Cref{tab:onedemo}.}
\label{tab:ours-limits}
\begin{tabular}{@{}ll@{}}
\toprule
Item & Limit \\
\midrule
Word--endpoint table & once per app, at most 8 entries \\
\textsc{Expect} & one expectation, at most 500 characters \\
\textsc{Suspect} & at most 5 suspicions, each at most 1{,}000 characters \\
\textsc{Test} & at most 50 read-only probes per verification; grounds at most 400 characters \\
Redrafts of a blocked cell & at most 3 per step \\
Decoding & temperature 0, at most 8{,}192 generated tokens per call \\
\bottomrule
\end{tabular}
\end{table}

\section{\ours{} prompts}
\label{app:prompts-ours}

\lstdefinestyle{prompt}{
  basicstyle=\ttfamily\scriptsize,
  columns=fullflexible,
  keepspaces=true,
  breaklines=true,
  breakatwhitespace=true,
  frame=single,
  framerule=0.4pt,
  xleftmargin=4pt,
  xrightmargin=4pt,
  aboveskip=6pt,
  belowskip=6pt,
  showstringspaces=false,
  upquote=true,
  inputencoding=utf8,
  extendedchars=true,
  literate={—}{{---}}1 {–}{{--}}1 {“}{{``}}1 {”}{{''}}1,
  captionpos=b,
}
\renewcommand{\lstlistingname}{Prompt}

\hcy{\ours{} runs the same stages on all three benchmarks. The prompt the agent starts from and the API it reads are
the benchmark's own. The verification prompts are AppWorld's on ALFWorld as well; on WebShop, \textsc{Suspect} and
\textsc{Test} also state how the shop works, \textsc{Suspect} words two of its five kinds for a shop, and \textsc{Test}
names the shop's read endpoints and defers two kinds of suspicion (Prompts~\ref{lst:ours-ws-suspect}
and~\ref{lst:ours-ws-test}). Values filled in at run time are shown in angle brackets; limits are those of the reported
runs.}

\subsection{AppWorld}
\label{app:prompts-ours-appworld}

\hcy{The agent drafts each action on AppWorld's official ReAct prompt with its one demonstration, to which it adds
the turns of Prompt~\ref{lst:ours-draft}. When it lists an app's endpoints, it answers
Prompt~\ref{lst:ours-ground} once for that app. Each drafted action that writes to the environment then goes
through \textsc{Expect} (Prompt~\ref{lst:ours-expect}), \textsc{Suspect} (Prompt~\ref{lst:ours-suspect}, or
Prompt~\ref{lst:ours-suspect-finish} when the action ends the task), and \textsc{Test}
(Prompts~\ref{lst:ours-test} and~\ref{lst:ours-test-followup}), and the agent's next turn begins with the verdicts
(Prompt~\ref{lst:ours-verdicts}).}

\begin{table}[H]
\centering
\small
\caption{The five kinds of falsifier the \textsc{Suspect} prompt names}
\label{tab:kinds}
\begin{tabular}{@{}lp{0.46\textwidth}l@{}}
\toprule
Kind & The falsifier asserts that & Derived from \\
\midrule
Wrong object   & the records acted on are not the ones the task names  & the action \\
Incomplete set & the task names more records than the action covers    & the action \\
Wrong value    & an argument or endpoint is not what the task means    & the action \\
Hidden state   & existing state changes what the call does            & the action \\
\midrule
The task's words & the part of the task this action carries out asks for something the action does not do &
  \begin{tabular}[t]{@{}l@{}}the task and\\ the word--endpoint table\end{tabular} \\
\bottomrule
\end{tabular}
\end{table}

\begin{lstlisting}[style=prompt,caption={Draft: the turns our agent adds to AppWorld's official ReAct prompt, after its first user turn. The rest of the prompt, including its one demonstration, is the official one; the demonstration's API documentation is rendered from the live environment.},label={lst:ours-draft}]
[[[ user ]]]
My name is: <first_name> <last_name>. My personal email is <email> and phone number is <phone_number>.
Task: How many playlists do I have in Spotify?

[[[ assistant ]]]
Let me first store your details and today's date in variables for later use.
Code:
```python
<code that stores these details and today's date in variables>
```

[[[ user ]]]
Output:
```
User: <first_name> <last_name> (<email>, <phone_number>); today: 2023-05-18 12:00:00
```
\end{lstlisting}

\begin{lstlisting}[style=prompt,caption={Word--endpoint table, asked once per app after the agent lists that app's endpoints.},label={lst:ours-ground}]
The list above is every endpoint of `<app>`. Before going on, match the task to it.

The task, verbatim:

> <task instruction>

For each thing the task asks this app to do, write one entry:

- `quote`: the task's words, copied word for word.
- `endpoint`: the endpoint whose description says it does that.
- `evidence`: the words of that description, copied exactly.
- `general`: if a second endpoint could also do it but its description is about a broader act or object, name it here; otherwise leave it empty. `endpoint` is then the one whose description names the same object or act as the quoted words.

At most 8 entries. Answer in exactly one fenced ```json``` block:

```json
[{"quote": "<words copied from the task>", "endpoint": "<endpoint name>", "evidence": "<words copied from that endpoint's description>", "general": "<the broader endpoint, or empty>"}, ...]
```
\end{lstlisting}

\begin{lstlisting}[style=prompt,caption={\textsc{Expect}.},label={lst:ours-expect}]
You proposed the code above. Before it executes, state clearly what you expect this code to accomplish for the user's task.

- Describe the resulting task-relevant state that should hold if this code succeeds. Do not describe your future plans.
- Be precise about which objects, values, or parts of the user's request this code is intended to affect. Do not claim that a broader requirement such as “all,” “each,” “only,” or “exactly” will be satisfied unless this single execution actually establishes it.

Answer in exactly one fenced ```json``` block, and keep it under 500 characters:

```json
{"expectation": "..."}
```
\end{lstlisting}

\begin{lstlisting}[style=prompt,caption={\textsc{Suspect}. The five kinds are those of \Cref{tab:kinds}; the table is the agent's answer to Prompt~\ref{lst:ours-ground}.},label={lst:ours-suspect}]
Before that cell runs, argue against yourself. What could make what you just said false when it actually runs? List the specific ways it could turn out wrong — not generic caution, and not doubts the environment rules out.

In the order it would run, that cell binds these variables and makes these API calls — every parameter of every call, including the optional ones it does not pass (`"passed": false`), which take the `default` shown:

```json
<the cell's variables and API calls, every parameter included>
```

Look for failures of these kinds:

1. **The wrong thing.** The objects this acts on may not be the ones the task asked for — a name that merely looks right, an id read off the wrong record.
2. **Not all of them.** The task may name a set or an amount this cell does not cover. If it acts on some members, say what would show there are others it missed.
3. **The wrong value or operation.** The arguments, quantities, or the endpoint itself may not be what the request means, even when the target is right.
4. **Hidden state.** Something already in the environment may change the outcome — what is in the cart, what the account already holds, what an earlier cell left behind.
5. **The task's words.** The part of the task this code carries out may ask for something the code does not do: an endpoint whose description matches those words more closely than the one it calls, or a condition, an amount or a second action in those words that the code leaves out.

When you read the apps' endpoints, you matched the task's words to these endpoints:

<the word--endpoint table>

Each entry is one way it could fail, in terms that a read of the environment could settle, and under 1000 characters. Say what it is about as its `ref`: one of the `ref` values listed above, exactly as written there. Give EVERY doubt you actually have, up to 5 — the number is not fixed, and the shape below is not a hint at how many. If you genuinely see none, answer with an empty list.

Answer in exactly one fenced ```json``` block:

```json
[{"doubt": "<one way it could turn out wrong>", "ref": "<one of the ref values listed above>"}, ...]
```
\end{lstlisting}

\begin{lstlisting}[style=prompt,caption={\textsc{Suspect} on a cell that ends the task.},label={lst:ours-suspect-finish}]
That cell ends the task. Before it runs, argue against yourself: what part of the task is not done yet?

This is the task:

<task instruction>

When you read the apps' endpoints, you matched these words of the task to endpoints that change state, and no cell above has called any of them:

<the task's words matched to state-changing endpoints that no cell has called>

Go through the task's words one by one — every action it asks for, every object it names, every condition or amount attached to them — and check each against the cells above and their Outputs. Look for:

1. **Not done.** An action the task asks for that no cell above carried out.
2. **Done in part.** An action carried out for only some of the objects the task names, or without a condition, amount or detail the task attaches to it.
3. **Done to the wrong thing.** An action carried out on objects other than the ones the task names.

Each entry is one part of the task that may not be done, in terms that a read of the environment could settle — what a read would show if it were not done — and under 1000 characters. Set its `ref` to `"task"`. Give EVERY such part you actually suspect, up to 5 — the number is not fixed. If every part of the task is done, answer with an empty list.

Answer in exactly one fenced ```json``` block:

```json
[{"doubt": "<one part of the task that may not be done>", "ref": "task"}, ...]
```
\end{lstlisting}

\begin{lstlisting}[style=prompt,caption={\textsc{Test}: the opening turn, stated once for all suspicions.},label={lst:ours-test}]
Wait. I have not run that code, because I'm suspicious about:

<the suspicions, numbered>

None of it has run, so none of the API calls in it have had any effect — what you read now is the world as it stands before that code.
---

Do your best to gather thorough and sufficient evidence to determine whether each suspicion should be confirmed or dismissed. Investigate them seriously in either direction: do not try to defend the proposed cell, but do not assume a suspicion is true either.

**Do not treat plausibility, possibility, or expectation as evidence.** When the current evidence does not decisively settle a suspicion, actively use evidence-gathering probes to obtain observations that can. Choose probes for their ability to distinguish the possibilities, and keep probing until you have concrete evidence or no useful probe remains.

Answer with exactly either a ```python``` or ```json``` block.

If more evidence is needed, return one ```python``` block containing an evidence-gathering probe:

```python
# Write a code to probe
```

To gather evidence, you may use any GET endpoints (including API documentation). If you want to access apps that are not currently logged in, you may perform the necessary `apis.<app>.login` action first. Reuse any tokens or variables already available from previous cells.

You can attempt <n> probes. Use them efficiently: prefer probes that are most likely to settle a suspicion and avoid redundant observations.

When you can judge, return one ```json``` block covering **<the suspicions' numbers>** — those are the ones an observation can settle, and the rest are not yours to decide here:

```json
{"<number>": {"verdict": "suspicion_is_confirmed" | "suspicion_is_dismissed", "grounds": "<concrete evidence, 400 characters at most>"}}
```

Use `"judgement_is_deferred"` for a suspicion no available evidence or probe can settle. If a useful probe could still distinguish the possibilities, probe instead of deferring.



A suspicion about the task's words — another endpoint, or a part of the task not done — is confirmed only by a line you quote as the grounds: an endpoint's description from the API docs, or an Output line showing the part is not done. A suspicion that only reads the task's words another way is `judgement_is_deferred`.
\end{lstlisting}

\begin{lstlisting}[style=prompt,caption={\textsc{Test}: follow-up turns, after a probe's output, for suspicions left without a verdict, and for a probe that would write.},label={lst:ours-test-followup}]


--- after a probe ---

Here is the output of your probe.

Use this new evidence to update your judgement.

Return exactly one fenced block:
- JSON if you can judge the suspicion.
- Python if another useful probe is needed.

--- suspicions left without a verdict ---

You did not give a verdict for these:

<the suspicions without a verdict>

Answer for them now, in one ```json``` block in the same shape — or probe again first if that is what they need. You can attempt <n> probes.

--- a probe that would write ---

I did not run that. It calls <the write endpoints it calls>, which the cell I am asking about does not call.

A probe only READS. The cell under suspicion has not run, so anything a probe changes changes the world for a cell that may still be rewritten — and a call that cell never makes is not evidence about that cell either.

Answer again: a ```json``` block if you can decide now, or a
```python``` block that only reads. <n> reads left.
\end{lstlisting}

\begin{lstlisting}[style=prompt,caption={The agent's next turn begins with the verdicts; the sentence ends one way when no suspicion is confirmed (the cell runs as written) and the other way otherwise (the agent writes a revised cell).},label={lst:ours-verdicts}]
<one verdict per suspicion, with its grounds>

So, nothing here stops it — running the cell as written:

--- otherwise ---

<one verdict per suspicion, with its grounds>

So,
\end{lstlisting}

\subsection{WebShop}
\label{app:prompts-ours-webshop}

\hcy{Thinking is off, and a task allows at most 50 steps. The agent starts from Prompt~\ref{lst:ours-ws-seed}: its
system turn is the one-sentence task definition of the default WebShop prompt of Harness-R1~\citep{harnessr1}, and its
user turns are a benchmark-neutral seed with no persona and no demonstration. It reads the shop's API
(Prompt~\ref{lst:ours-ws-api}) through \code{api\_docs}. A task ends only with a purchase: before \code{buy}, a call to
\code{complete\_task} is refused with ``This shopping session ends only with a purchase: call
apis.webshop.buy.'' \textsc{Suspect} and \textsc{Test} state how the shop works (Prompts~\ref{lst:ours-ws-suspect}
and~\ref{lst:ours-ws-test}). The word--endpoint table, \textsc{Expect}, \textsc{Suspect} on a cell that ends the task,
and the later turns of \textsc{Test} use the AppWorld prompts (Prompts~\ref{lst:ours-ground}, \ref{lst:ours-expect},
\ref{lst:ours-suspect-finish}, \ref{lst:ours-test-followup}, and~\ref{lst:ours-verdicts}).}

\begin{lstlisting}[style=prompt,caption={WebShop: the first turns. The system turn is the task definition of Harness-R1's default WebShop prompt; the user turns are our benchmark-neutral seed. In the first user turn, \code{<app>}, \code{<api>}, and \code{<answer>} are part of the prompt itself.},label={lst:ours-ws-seed}]
[[[ system ]]]
You are a web shopping agent. Follow the task instruction to find and buy the correct product.

[[[ user ]]]
You are a super intelligent AI Assistant whose job is to achieve the given task completely autonomously. The shopping instructions above govern everything you do; where they and this message disagree, they win.

To do this, you will interact with a set of apps through their associated APIs. For this you will undertake a *multi-step conversation* using a python REPL environment. That is, you will write python code and the environment will execute it and show you the result, based on which you will write python code for the next step, and so on, until you've achieved the goal.

Here are three key APIs that you need to know to get more information

# To get the list of apps that are available to you.

```python
print(apis.api_docs.show_app_descriptions())
```

# To get the list of apis under any app listed above, e.g. an app named `<app>`

```python
print(apis.api_docs.show_api_descriptions(app_name='<app>'))
```

# To get the specification of a particular api, e.g. app `<app>`'s api `<api>`

```python
print(apis.api_docs.show_api_doc(app_name='<app>', api_name='<api>'))
```

Each code execution will produce an output that you can use in subsequent calls. Using these APIs, you can now generate code, that I will execute, to solve the task.

The apps currently available to you are:
[
 {
  "description": "An online shop. Search its catalogue, open a product and read its sections, choose the product\u0027s options, and buy it. One purchase per session; buying ends the session.",
  "name": "webshop"
 },
 {
  "description": "Task-lifecycle APIs (mark the task complete).",
  "name": "supervisor"
 }
]

**Key instructions**:
(1) Make sure to end code blocks with ``` followed by a newline(\n).

(2) Remember you can use the variables in your code in subsequent code blocks.

(3) Always look at API specifications (using apis.api_docs.show_api_doc) before calling an API.

(4) Write small chunks of code and only one chunk of code in every step. Make sure everything is working correctly before making any irreversible change.

(5) Once you have completed the task, make sure to call apis.supervisor.complete_task(). If the task asked for some information, return it as the answer argument, i.e. call apis.supervisor.complete_task(answer=<answer>). Many tasks do not require an answer, so in those cases, just call apis.supervisor.complete_task() i.e. do not pass any argument.


[[[ assistant ]]]
Understood. I'll consult the app and API docs as needed, work in small verified steps, and call `apis.supervisor.complete_task()` when done. What is the task?

[[[ user ]]]
Using these APIs, generate code to solve the actual task:

Task: <task instruction>
\end{lstlisting}

\begin{lstlisting}[style=prompt,caption={WebShop: the shop's API documentation as it enters the agent's context. Each \code{api\_docs} cell is followed by its Output, verbatim from the runs: the list of the shop's endpoints, then the specification of each.},label={lst:ours-ws-api}]
[[[ assistant ]]]
```python
print(apis.api_docs.show_api_descriptions(app_name='webshop'))
```

[[[ user ]]]
Output:
```
INFO apis.api_docs.show_api_descriptions {"app_name": "webshop"} -> [{"name": "search", "description": "Search the catalogue. Returns one page of ten results (asin, title, price) and the total number of results."}, {"name": "view_product", "description": "Open a product from your search results: its title, price, rating, the options it can be bought with, and the options selected so far."}, {"name": "product_section", "description": "Read one section of a product from your search results: `Description`, `Features` or `Reviews`."}, {"name": "select_option", "description": "Choose one value of one of a product's options, e.g. its size or color. Choosing again replaces the earlier value."}, {"name": "buy", "description": "Buy a product with the options selected for it. Ends the session; it cannot be undone."}]
[
 {
  "name": "search",
  "description": "Search the catalogue. Returns one page of ten results (asin, title, price) and the total number of results."
 },
 {
  "name": "view_product",
  "description": "Open a product from your search results: its title, price, rating, the options it can be bought with, and the options selected so far."
 },
 {
  "name": "product_section",
  "description": "Read one section of a product from your search results: `Description`, `Features` or `Reviews`."
 },
 {
  "name": "select_option",
  "description": "Choose one value of one of a product's options, e.g. its size or color. Choosing again replaces the earlier value."
 },
 {
  "name": "buy",
  "description": "Buy a product with the options selected for it. Ends the session; it cannot be undone."
 }
]
```

[[[ assistant ]]]
```python
print(apis.api_docs.show_api_doc(app_name='webshop', api_name='search'))
```

[[[ user ]]]
Output:
```
INFO apis.api_docs.show_api_doc {"app_name": "webshop", "api_name": "search"} -> {"app_name": "webshop", "api_name": "search", "path": "/webshop/search", "method": "GET", "description": "Search the catalogue. Returns one page of ten results (asin, title, price) and the total number of results.", "parameters": [{"name": "query", "required": true, "type": "string"}, {"name": "page", "required": false, "type": "integer"}], "response_schemas": {"success": {"page": "any", "total": "any", "results": "any"}, "failure": "an exception is raised"}}
{
 "app_name": "webshop",
 "api_name": "search",
 "path": "/webshop/search",
 "method": "GET",
 "description": "Search the catalogue. Returns one page of ten results (asin, title, price) and the total number of results.",
 "parameters": [
  {
   "name": "query",
   "required": true,
   "type": "string"
  },
  {
   "name": "page",
   "required": false,
   "type": "integer"
  }
 ],
 "response_schemas": {
  "success": {
   "page": "any",
   "total": "any",
   "results": "any"
  },
  "failure": "an exception is raised"
 }
}
```

[[[ assistant ]]]
```python
print(apis.api_docs.show_api_doc(app_name='webshop', api_name='view_product'))
```

[[[ user ]]]
Output:
```
INFO apis.api_docs.show_api_doc {"app_name": "webshop", "api_name": "view_product"} -> {"app_name": "webshop", "api_name": "view_product", "path": "/webshop/view_product", "method": "GET", "description": "Open a product from your search results: its title, price, rating, the options it can be bought with, and the options selected so far.", "parameters": [{"name": "asin", "required": true, "type": "string"}], "response_schemas": {"success": {"asin": "any", "title": "any", "price": "any", "rating": "any", "options": "any", "selected_options": "any"}, "failure": "an exception is raised"}}
{
 "app_name": "webshop",
 "api_name": "view_product",
 "path": "/webshop/view_product",
 "method": "GET",
 "description": "Open a product from your search results: its title, price, rating, the options it can be bought with, and the options selected so far.",
 "parameters": [
  {
   "name": "asin",
   "required": true,
   "type": "string"
  }
 ],
 "response_schemas": {
  "success": {
   "asin": "any",
   "title": "any",
   "price": "any",
   "rating": "any",
   "options": "any",
   "selected_options": "any"
  },
  "failure": "an exception is raised"
 }
}
```

[[[ assistant ]]]
```python
print(apis.api_docs.show_api_doc(app_name='webshop', api_name='product_section'))
```

[[[ user ]]]
Output:
```
INFO apis.api_docs.show_api_doc {"app_name": "webshop", "api_name": "product_section"} -> {"app_name": "webshop", "api_name": "product_section", "path": "/webshop/product_section", "method": "GET", "description": "Read one section of a product from your search results: `Description`, `Features` or `Reviews`.", "parameters": [{"name": "asin", "required": true, "type": "string"}, {"name": "section", "required": true, "type": "string"}], "response_schemas": {"success": {"asin": "any", "section": "any", "content": "any"}, "failure": "an exception is raised"}}
{
 "app_name": "webshop",
 "api_name": "product_section",
 "path": "/webshop/product_section",
 "method": "GET",
 "description": "Read one section of a product from your search results: `Description`, `Features` or `Reviews`.",
 "parameters": [
  {
   "name": "asin",
   "required": true,
   "type": "string"
  },
  {
   "name": "section",
   "required": true,
   "type": "string"
  }
 ],
 "response_schemas": {
  "success": {
   "asin": "any",
   "section": "any",
   "content": "any"
  },
  "failure": "an exception is raised"
 }
}
```

[[[ assistant ]]]
```python
print(apis.api_docs.show_api_doc(app_name='webshop', api_name='select_option'))
```

[[[ user ]]]
Output:
```
INFO apis.api_docs.show_api_doc {"app_name": "webshop", "api_name": "select_option"} -> {"app_name": "webshop", "api_name": "select_option", "path": "/webshop/select_option", "method": "POST", "description": "Choose one value of one of a product's options, e.g. its size or color. Choosing again replaces the earlier value.", "parameters": [{"name": "asin", "required": true, "type": "string"}, {"name": "option_type", "required": true, "type": "string"}, {"name": "value", "required": true, "type": "string"}], "response_schemas": {"success": {"asin": "any", "selected_options": "any"}, "failure": "an exception is raised"}}
{
 "app_name": "webshop",
 "api_name": "select_option",
 "path": "/webshop/select_option",
 "method": "POST",
 "description": "Choose one value of one of a product's options, e.g. its size or color. Choosing again replaces the earlier value.",
 "parameters": [
  {
   "name": "asin",
   "required": true,
   "type": "string"
  },
  {
   "name": "option_type",
   "required": true,
   "type": "string"
  },
  {
   "name": "value",
   "required": true,
   "type": "string"
  }
 ],
 "response_schemas": {
  "success": {
   "asin": "any",
   "selected_options": "any"
  },
  "failure": "an exception is raised"
 }
}
```

[[[ assistant ]]]
```python
print(apis.api_docs.show_api_doc(app_name='webshop', api_name='buy'))
```

[[[ user ]]]
Output:
```
INFO apis.api_docs.show_api_doc {"app_name": "webshop", "api_name": "buy"} -> {"app_name": "webshop", "api_name": "buy", "path": "/webshop/buy", "method": "POST", "description": "Buy a product with the options selected for it. Ends the session; it cannot be undone.", "parameters": [{"name": "asin", "required": true, "type": "string"}], "response_schemas": {"success": {"message": "any", "asin": "any", "selected_options": "any"}, "failure": "an exception is raised"}}
{
 "app_name": "webshop",
 "api_name": "buy",
 "path": "/webshop/buy",
 "method": "POST",
 "description": "Buy a product with the options selected for it. Ends the session; it cannot be undone.",
 "parameters": [
  {
   "name": "asin",
   "required": true,
   "type": "string"
  }
 ],
 "response_schemas": {
  "success": {
   "message": "any",
   "asin": "any",
   "selected_options": "any"
  },
  "failure": "an exception is raised"
 }
}
```

[[[ assistant ]]]
```python
print(apis.api_docs.show_api_doc(app_name='supervisor', api_name='complete_task'))
```

[[[ user ]]]
Output:
```
INFO apis.api_docs.show_api_doc {"app_name": "supervisor", "api_name": "complete_task"} -> {"app_name": "supervisor", "api_name": "complete_task", "path": "/supervisor/complete_task", "method": "POST", "description": "Signal that the task is finished. Call once, last, with the answer if the task asks a question.", "parameters": [{"name": "answer", "required": false, "type": "string"}], "response_schemas": {"success": {"message": "string"}, "failure": "an exception is raised"}}
{
 "app_name": "supervisor",
 "api_name": "complete_task",
 "path": "/supervisor/complete_task",
 "method": "POST",
 "description": "Signal that the task is finished. Call once, last, with the answer if the task asks a question.",
 "parameters": [
  {
   "name": "answer",
   "required": false,
   "type": "string"
  }
 ],
 "response_schemas": {
  "success": {
   "message": "string"
  },
  "failure": "an exception is raised"
 }
}
```
\end{lstlisting}

\begin{lstlisting}[style=prompt,caption={WebShop: \textsc{Suspect}. It is Prompt~\ref{lst:ours-suspect} with the block \emph{About this shop} added and kinds 2 and 4 worded for a shop.},label={lst:ours-ws-suspect}]
Before that cell runs, argue against yourself. What could make what you just said false when it actually runs? List the specific ways it could turn out wrong — not generic caution, and not doubts the environment rules out.

In the order it would run, that cell binds these variables and makes these API calls — every parameter of every call, including the optional ones it does not pass (`"passed": false`), which take the `default` shown:

```json
<the cell's variables and API calls, every parameter included>
```

About this shop — how it works, not a hint at any answer:
- A task here is completed by buying exactly one product with `apis.webshop.buy`. Words such as "find", "looking for", "need" or "search" in the task say which product to buy; they do not ask for a report instead of a purchase.
- `apis.webshop.select_option` sets one option of one product per call. A cell that selects some of a product's options is one step: whether every option the task names is selected, and whether the product is bought, is settled at the cell that calls `apis.webshop.buy`, not at this one.

Look for failures of these kinds:

1. **The wrong thing.** The objects this acts on may not be the ones the task asked for — a name that merely looks right, an id read off the wrong record.
2. **Not all of them.** At the cell that buys: the task may name options or an amount the product as bought does not have — say which, and what a read would show.
3. **The wrong value or operation.** The arguments, quantities, or the endpoint itself may not be what the request means, even when the target is right.
4. **Hidden state.** Something already in the shop may change the outcome — the options already selected for this product, what an earlier cell left behind.
5. **The task's words.** The part of the task this code carries out may ask for something the code does not do: an endpoint whose description matches those words more closely than the one it calls, or a condition, an amount or a second action in those words that the code leaves out.

When you read the apps' endpoints, you matched the task's words to these endpoints:

<the word--endpoint table>

Each entry is one way it could fail, in terms that a read of the environment could settle, and under 1000 characters. Say what it is about as its `ref`: one of the `ref` values listed above, exactly as written there. Give EVERY doubt you actually have, up to 5 — the number is not fixed, and the shape below is not a hint at how many. If you genuinely see none, answer with an empty list.

Answer in exactly one fenced ```json``` block:

```json
[{"doubt": "<one way it could turn out wrong>", "ref": "<one of the ref values listed above>"}, ...]
```
\end{lstlisting}

\begin{lstlisting}[style=prompt,caption={WebShop: \textsc{Test}, opening turn. It is Prompt~\ref{lst:ours-test} with the shop's read endpoints named in place of AppWorld's login step, the block \emph{About this shop}, and the last paragraph added.},label={lst:ours-ws-test}]
Wait. I have not run that code, because I'm suspicious about:

<the suspicions, numbered>

None of it has run, so none of the API calls in it have had any effect — what you read now is the world as it stands before that code.
---

Do your best to gather thorough and sufficient evidence to determine whether each suspicion should be confirmed or dismissed. Investigate them seriously in either direction: do not try to defend the proposed cell, but do not assume a suspicion is true either.

**Do not treat plausibility, possibility, or expectation as evidence.** When the current evidence does not decisively settle a suspicion, actively use evidence-gathering probes to obtain observations that can. Choose probes for their ability to distinguish the possibilities, and keep probing until you have concrete evidence or no useful probe remains.

Answer with exactly either a ```python``` or ```json``` block.

If more evidence is needed, return one ```python``` block containing an evidence-gathering probe:

```python
# Write a code to probe
```

To gather evidence, you may use the shop's read endpoints — `apis.webshop.search`, `apis.webshop.view_product`, `apis.webshop.product_section` — and the API documentation. Reuse any variables already available from previous cells.

About this shop — how it works, not a hint at any answer:
- A task here is completed by buying exactly one product with `apis.webshop.buy`. Words such as "find", "looking for", "need" or "search" in the task say which product to buy; they do not ask for a report instead of a purchase.
- `apis.webshop.select_option` sets one option of one product per call. A cell that selects some of a product's options is one step: whether every option the task names is selected, and whether the product is bought, is settled at the cell that calls `apis.webshop.buy`, not at this one.
- Every task in this shop was written for a product in its catalogue.

You can attempt <n> probes. Use them efficiently: prefer probes that are most likely to settle a suspicion and avoid redundant observations.

When you can judge, return one ```json``` block covering **<the suspicions' numbers>** — those are the ones an observation can settle, and the rest are not yours to decide here:

```json
{"<number>": {"verdict": "suspicion_is_confirmed" | "suspicion_is_dismissed", "grounds": "<concrete evidence, 400 characters at most>"}}
```

Use `"judgement_is_deferred"` for a suspicion no available evidence or probe can settle. If a useful probe could still distinguish the possibilities, probe instead of deferring.



A suspicion about the task's words — another endpoint, or a part of the task not done — is confirmed only by a line you quote as the grounds: an endpoint's description from the API docs, or an Output line showing the part is not done. A suspicion that only reads the task's words another way is `judgement_is_deferred`.

A suspicion that a `select_option` cell leaves other options unselected, or that it does not buy, is not about that cell: judge it `judgement_is_deferred`. A suspicion that no product in the shop fits the task is not confirmed by searches that did not find one: judge it `judgement_is_deferred`.
\end{lstlisting}

\subsection{ALFWorld}
\label{app:prompts-ours-alfworld}

\hcy{Thinking is on. The agent starts from Prompt~\ref{lst:ours-alf-seed}: a benchmark-neutral seed with no persona,
then the two demonstrations that ReAct~\citep{react} shows for the task's type, verbatim, and the game's opening text.
It acts through one app whose endpoints are the game's commands (Prompt~\ref{lst:ours-alf-api}). \code{look},
\code{inventory}, and \code{examine} are reads: a cell that calls only these runs without verification, and they are
the commands a probe may send. An episode allows 50 game commands, ALFWorld's standard limit, and every command counts,
probes included. All verification prompts are the AppWorld ones (Prompts~\ref{lst:ours-ground}
to~\ref{lst:ours-verdicts}).}

\begin{lstlisting}[style=prompt,caption={ALFWorld: the first turns, our benchmark-neutral seed with the game's two ReAct demonstrations for the task's type appended to the first turn, verbatim. In the first user turn, \code{<app>}, \code{<api>}, and \code{<answer>} are part of the prompt itself.},label={lst:ours-alf-seed}]
[[[ user ]]]
You are a super intelligent AI Assistant whose job is to achieve the given task completely autonomously.

To do this, you will interact with a set of apps through their associated APIs. For this you will undertake a *multi-step conversation* using a python REPL environment. That is, you will write python code and the environment will execute it and show you the result, based on which you will write python code for the next step, and so on, until you've achieved the goal.

Here are three key APIs that you need to know to get more information

# To get the list of apps that are available to you.

```python
print(apis.api_docs.show_app_descriptions())
```

# To get the list of apis under any app listed above, e.g. an app named `<app>`

```python
print(apis.api_docs.show_api_descriptions(app_name='<app>'))
```

# To get the specification of a particular api, e.g. app `<app>`'s api `<api>`

```python
print(apis.api_docs.show_api_doc(app_name='<app>', api_name='<api>'))
```

Each code execution will produce an output that you can use in subsequent calls. Using these APIs, you can now generate code, that I will execute, to solve the task.

The apps currently available to you are:
[
 {
  "description": "The household you are in. Look around, check what you carry, examine things, move between receptacles, open and close them, take and put objects, and heat, cool, clean or switch on things. Objects and receptacles are named by type and index (`mug 1`, `cabinet 3`); a type name is exact \u2014 a similarly named type is a different type. The task is judged on the household\u0027s state when you finish: where every object is, in your hands included, and what has been done to it (heated, cooled, cleaned, switched on).",
  "name": "alfworld"
 },
 {
  "description": "Task-lifecycle APIs (mark the task complete).",
  "name": "supervisor"
 }
]

**Key instructions**:
(1) Make sure to end code blocks with ``` followed by a newline(\n).

(2) Remember you can use the variables in your code in subsequent code blocks.

(3) Always look at API specifications (using apis.api_docs.show_api_doc) before calling an API.

(4) Write small chunks of code and only one chunk of code in every step. Make sure everything is working correctly before making any irreversible change.

(5) Once you have completed the task, make sure to call apis.supervisor.complete_task(). If the task asked for some information, return it as the answer argument, i.e. call apis.supervisor.complete_task(answer=<answer>). Many tasks do not require an answer, so in those cases, just call apis.supervisor.complete_task() i.e. do not pass any argument.


Here are 2 example episodes of this household game as its players see it — `>` lines are the player's commands, the lines after each are the game's reply, and `think:` lines are the player's own notes; each command corresponds to one API of the app above.

```
<ReAct's first demonstration for the task's type>
```
```
<ReAct's second demonstration for the task's type>
```


[[[ assistant ]]]
Understood. I'll consult the app and API docs as needed, work in small verified steps, and call `apis.supervisor.complete_task()` when done. What is the task?

[[[ user ]]]
Using these APIs, generate code to solve the actual task.

<the game's opening text: the room and the task>

Task: <task instruction>
\end{lstlisting}

\begin{lstlisting}[style=prompt,caption={ALFWorld: the game's API documentation as it enters the agent's context. Each \code{api\_docs} cell is followed by its Output, verbatim from the runs: the list of the game's endpoints, one per command, then the specification of each. The agent never read the specification of \code{close} in these runs; it is shown as the environment returns it.},label={lst:ours-alf-api}]
[[[ assistant ]]]
```python
print(apis.api_docs.show_api_descriptions(app_name='alfworld'))
```

[[[ user ]]]
Output:
```
INFO apis.api_docs.show_api_descriptions {"app_name": "alfworld"} -> [{"name": "look", "description": "Describe your current location and what is visible there."}, {"name": "inventory", "description": "List what you are carrying."}, {"name": "examine", "description": "Look closely at an object or a receptacle you are at \u2014 a description, nothing more; for a receptacle it repeats what arriving there showed."}, {"name": "go_to", "description": "Move to a receptacle, e.g. `countertop 1`; on arrival you are told what is on or in it (a closed one must be opened first)."}, {"name": "open", "description": "Open a closed receptacle you are at."}, {"name": "close", "description": "Close an open receptacle you are at."}, {"name": "take", "description": "Pick up an object from the receptacle you are at. You can carry one object at a time: put it down before taking another."}, {"name": "put", "description": "Put an object you carry in or on the receptacle you are at."}, {"name": "heat", "description": "Heat an object you carry with an appliance you are at."}, {"name": "cool", "description": "Cool an object you carry with an appliance you are at."}, {"name": "clean", "description": "Clean an object you carry with a basin you are at."}, {"name": "use", "description": "Switch on an object that can be turned on (a lamp, for instance), at the receptacle you are at."}]
[
 {
  "name": "look",
  "description": "Describe your current location and what is visible there."
 },
 {
  "name": "inventory",
  "description": "List what you are carrying."
 },
 {
  "name": "examine",
  "description": "Look closely at an object or a receptacle you are at \u2014 a description, nothing more; for a receptacle it repeats what arriving there showed."
 },
 {
  "name": "go_to",
  "description": "Move to a receptacle, e.g. `countertop 1`; on arrival you are told what is on or in it (a closed one must be opened first)."
 },
 {
  "name": "open",
  "description": "Open a closed receptacle you are at."
 },
 {
  "name": "close",
  "description": "Close an open receptacle you are at."
 },
 {
  "name": "take",
  "description": "Pick up an object from the receptacle you are at. You can carry one object at a time: put it down before taking another."
 },
 {
  "name": "put",
  "description": "Put an object you carry in or on the receptacle you are at."
 },
 {
  "name": "heat",
  "description": "Heat an object you carry with an appliance you are at."
 },
 {
  "name": "cool",
  "description": "Cool an object you carry with an appliance you are at."
 },
 {
  "name": "clean",
  "description": "Clean an object you carry with a basin you are at."
 },
 {
  "name": "use",
  "description": "Switch on an object that can be turned on (a lamp, for instance), at the receptacle you are at."
 }
]
```

[[[ assistant ]]]
```python
print(apis.api_docs.show_api_doc(app_name='alfworld', api_name='look'))
```

[[[ user ]]]
Output:
```
INFO apis.api_docs.show_api_doc {"app_name": "alfworld", "api_name": "look"} -> {"app_name": "alfworld", "api_name": "look", "path": "/alfworld/look", "method": "GET", "description": "Describe your current location and what is visible there.", "parameters": [], "response_schemas": {"success": {"observation": "string"}, "failure": "an exception is raised"}}
{
 "app_name": "alfworld",
 "api_name": "look",
 "path": "/alfworld/look",
 "method": "GET",
 "description": "Describe your current location and what is visible there.",
 "parameters": [],
 "response_schemas": {
  "success": {
   "observation": "string"
  },
  "failure": "an exception is raised"
 }
}
```

[[[ assistant ]]]
```python
print(apis.api_docs.show_api_doc(app_name='alfworld', api_name='inventory'))
```

[[[ user ]]]
Output:
```
INFO apis.api_docs.show_api_doc {"app_name": "alfworld", "api_name": "inventory"} -> {"app_name": "alfworld", "api_name": "inventory", "path": "/alfworld/inventory", "method": "GET", "description": "List what you are carrying.", "parameters": [], "response_schemas": {"success": {"observation": "string"}, "failure": "an exception is raised"}}
{
 "app_name": "alfworld",
 "api_name": "inventory",
 "path": "/alfworld/inventory",
 "method": "GET",
 "description": "List what you are carrying.",
 "parameters": [],
 "response_schemas": {
  "success": {
   "observation": "string"
  },
  "failure": "an exception is raised"
 }
}
```

[[[ assistant ]]]
```python
print(apis.api_docs.show_api_doc(app_name='alfworld', api_name='examine'))
```

[[[ user ]]]
Output:
```
INFO apis.api_docs.show_api_doc {"app_name": "alfworld", "api_name": "examine"} -> {"app_name": "alfworld", "api_name": "examine", "path": "/alfworld/examine", "method": "GET", "description": "Look closely at an object or a receptacle you are at \u2014 a description, nothing more; for a receptacle it repeats what arriving there showed.", "parameters": [{"name": "object", "required": true, "type": "string"}], "response_schemas": {"success": {"observation": "string"}, "failure": "an exception is raised"}}
{
 "app_name": "alfworld",
 "api_name": "examine",
 "path": "/alfworld/examine",
 "method": "GET",
 "description": "Look closely at an object or a receptacle you are at \u2014 a description, nothing more; for a receptacle it repeats what arriving there showed.",
 "parameters": [
  {
   "name": "object",
   "required": true,
   "type": "string"
  }
 ],
 "response_schemas": {
  "success": {
   "observation": "string"
  },
  "failure": "an exception is raised"
 }
}
```

[[[ assistant ]]]
```python
print(apis.api_docs.show_api_doc(app_name='alfworld', api_name='go_to'))
```

[[[ user ]]]
Output:
```
INFO apis.api_docs.show_api_doc {"app_name": "alfworld", "api_name": "go_to"} -> {"app_name": "alfworld", "api_name": "go_to", "path": "/alfworld/go_to", "method": "POST", "description": "Move to a receptacle, e.g. `countertop 1`; on arrival you are told what is on or in it (a closed one must be opened first).", "parameters": [{"name": "receptacle", "required": true, "type": "string"}], "response_schemas": {"success": {"observation": "string"}, "failure": "an exception is raised"}}
{
 "app_name": "alfworld",
 "api_name": "go_to",
 "path": "/alfworld/go_to",
 "method": "POST",
 "description": "Move to a receptacle, e.g. `countertop 1`; on arrival you are told what is on or in it (a closed one must be opened first).",
 "parameters": [
  {
   "name": "receptacle",
   "required": true,
   "type": "string"
  }
 ],
 "response_schemas": {
  "success": {
   "observation": "string"
  },
  "failure": "an exception is raised"
 }
}
```

[[[ assistant ]]]
```python
print(apis.api_docs.show_api_doc(app_name='alfworld', api_name='open'))
```

[[[ user ]]]
Output:
```
INFO apis.api_docs.show_api_doc {"app_name": "alfworld", "api_name": "open"} -> {"app_name": "alfworld", "api_name": "open", "path": "/alfworld/open", "method": "POST", "description": "Open a closed receptacle you are at.", "parameters": [{"name": "receptacle", "required": true, "type": "string"}], "response_schemas": {"success": {"observation": "string"}, "failure": "an exception is raised"}}
{
 "app_name": "alfworld",
 "api_name": "open",
 "path": "/alfworld/open",
 "method": "POST",
 "description": "Open a closed receptacle you are at.",
 "parameters": [
  {
   "name": "receptacle",
   "required": true,
   "type": "string"
  }
 ],
 "response_schemas": {
  "success": {
   "observation": "string"
  },
  "failure": "an exception is raised"
 }
}
```

[[[ assistant ]]]
```python
print(apis.api_docs.show_api_doc(app_name='alfworld', api_name='close'))
```

[[[ user ]]]
Output:
```
INFO apis.api_docs.show_api_doc {"app_name": "alfworld", "api_name": "close"} -> {"app_name": "alfworld", "api_name": "close", "path": "/alfworld/close", "method": "POST", "description": "Close an open receptacle you are at.", "parameters": [{"name": "receptacle", "required": true, "type": "string"}], "response_schemas": {"success": {"observation": "string"}, "failure": "an exception is raised"}}
{
 "app_name": "alfworld",
 "api_name": "close",
 "path": "/alfworld/close",
 "method": "POST",
 "description": "Close an open receptacle you are at.",
 "parameters": [
  {
   "name": "receptacle",
   "required": true,
   "type": "string"
  }
 ],
 "response_schemas": {
  "success": {
   "observation": "string"
  },
  "failure": "an exception is raised"
 }
}
```

[[[ assistant ]]]
```python
print(apis.api_docs.show_api_doc(app_name='alfworld', api_name='take'))
```

[[[ user ]]]
Output:
```
INFO apis.api_docs.show_api_doc {"app_name": "alfworld", "api_name": "take"} -> {"app_name": "alfworld", "api_name": "take", "path": "/alfworld/take", "method": "POST", "description": "Pick up an object from the receptacle you are at. You can carry one object at a time: put it down before taking another.", "parameters": [{"name": "object", "required": true, "type": "string"}, {"name": "receptacle", "required": true, "type": "string"}], "response_schemas": {"success": {"observation": "string"}, "failure": "an exception is raised"}}
{
 "app_name": "alfworld",
 "api_name": "take",
 "path": "/alfworld/take",
 "method": "POST",
 "description": "Pick up an object from the receptacle you are at. You can carry one object at a time: put it down before taking another.",
 "parameters": [
  {
   "name": "object",
   "required": true,
   "type": "string"
  },
  {
   "name": "receptacle",
   "required": true,
   "type": "string"
  }
 ],
 "response_schemas": {
  "success": {
   "observation": "string"
  },
  "failure": "an exception is raised"
 }
}
```

[[[ assistant ]]]
```python
print(apis.api_docs.show_api_doc(app_name='alfworld', api_name='put'))
```

[[[ user ]]]
Output:
```
INFO apis.api_docs.show_api_doc {"app_name": "alfworld", "api_name": "put"} -> {"app_name": "alfworld", "api_name": "put", "path": "/alfworld/put", "method": "POST", "description": "Put an object you carry in or on the receptacle you are at.", "parameters": [{"name": "object", "required": true, "type": "string"}, {"name": "receptacle", "required": true, "type": "string"}], "response_schemas": {"success": {"observation": "string"}, "failure": "an exception is raised"}}
{
 "app_name": "alfworld",
 "api_name": "put",
 "path": "/alfworld/put",
 "method": "POST",
 "description": "Put an object you carry in or on the receptacle you are at.",
 "parameters": [
  {
   "name": "object",
   "required": true,
   "type": "string"
  },
  {
   "name": "receptacle",
   "required": true,
   "type": "string"
  }
 ],
 "response_schemas": {
  "success": {
   "observation": "string"
  },
  "failure": "an exception is raised"
 }
}
```

[[[ assistant ]]]
```python
print(apis.api_docs.show_api_doc(app_name='alfworld', api_name='heat'))
```

[[[ user ]]]
Output:
```
INFO apis.api_docs.show_api_doc {"app_name": "alfworld", "api_name": "heat"} -> {"app_name": "alfworld", "api_name": "heat", "path": "/alfworld/heat", "method": "POST", "description": "Heat an object you carry with an appliance you are at.", "parameters": [{"name": "object", "required": true, "type": "string"}, {"name": "receptacle", "required": true, "type": "string"}], "response_schemas": {"success": {"observation": "string"}, "failure": "an exception is raised"}}
{
 "app_name": "alfworld",
 "api_name": "heat",
 "path": "/alfworld/heat",
 "method": "POST",
 "description": "Heat an object you carry with an appliance you are at.",
 "parameters": [
  {
   "name": "object",
   "required": true,
   "type": "string"
  },
  {
   "name": "receptacle",
   "required": true,
   "type": "string"
  }
 ],
 "response_schemas": {
  "success": {
   "observation": "string"
  },
  "failure": "an exception is raised"
 }
}
```

[[[ assistant ]]]
```python
print(apis.api_docs.show_api_doc(app_name='alfworld', api_name='cool'))
```

[[[ user ]]]
Output:
```
INFO apis.api_docs.show_api_doc {"app_name": "alfworld", "api_name": "cool"} -> {"app_name": "alfworld", "api_name": "cool", "path": "/alfworld/cool", "method": "POST", "description": "Cool an object you carry with an appliance you are at.", "parameters": [{"name": "object", "required": true, "type": "string"}, {"name": "receptacle", "required": true, "type": "string"}], "response_schemas": {"success": {"observation": "string"}, "failure": "an exception is raised"}}
{
 "app_name": "alfworld",
 "api_name": "cool",
 "path": "/alfworld/cool",
 "method": "POST",
 "description": "Cool an object you carry with an appliance you are at.",
 "parameters": [
  {
   "name": "object",
   "required": true,
   "type": "string"
  },
  {
   "name": "receptacle",
   "required": true,
   "type": "string"
  }
 ],
 "response_schemas": {
  "success": {
   "observation": "string"
  },
  "failure": "an exception is raised"
 }
}
```

[[[ assistant ]]]
```python
print(apis.api_docs.show_api_doc(app_name='alfworld', api_name='clean'))
```

[[[ user ]]]
Output:
```
INFO apis.api_docs.show_api_doc {"app_name": "alfworld", "api_name": "clean"} -> {"app_name": "alfworld", "api_name": "clean", "path": "/alfworld/clean", "method": "POST", "description": "Clean an object you carry with a basin you are at.", "parameters": [{"name": "object", "required": true, "type": "string"}, {"name": "receptacle", "required": true, "type": "string"}], "response_schemas": {"success": {"observation": "string"}, "failure": "an exception is raised"}}
{
 "app_name": "alfworld",
 "api_name": "clean",
 "path": "/alfworld/clean",
 "method": "POST",
 "description": "Clean an object you carry with a basin you are at.",
 "parameters": [
  {
   "name": "object",
   "required": true,
   "type": "string"
  },
  {
   "name": "receptacle",
   "required": true,
   "type": "string"
  }
 ],
 "response_schemas": {
  "success": {
   "observation": "string"
  },
  "failure": "an exception is raised"
 }
}
```

[[[ assistant ]]]
```python
print(apis.api_docs.show_api_doc(app_name='alfworld', api_name='use'))
```

[[[ user ]]]
Output:
```
INFO apis.api_docs.show_api_doc {"app_name": "alfworld", "api_name": "use"} -> {"app_name": "alfworld", "api_name": "use", "path": "/alfworld/use", "method": "POST", "description": "Switch on an object that can be turned on (a lamp, for instance), at the receptacle you are at.", "parameters": [{"name": "object", "required": true, "type": "string"}], "response_schemas": {"success": {"observation": "string"}, "failure": "an exception is raised"}}
{
 "app_name": "alfworld",
 "api_name": "use",
 "path": "/alfworld/use",
 "method": "POST",
 "description": "Switch on an object that can be turned on (a lamp, for instance), at the receptacle you are at.",
 "parameters": [
  {
   "name": "object",
   "required": true,
   "type": "string"
  }
 ],
 "response_schemas": {
  "success": {
   "observation": "string"
  },
  "failure": "an exception is raised"
 }
}
```

[[[ assistant ]]]
```python
print(apis.api_docs.show_api_doc(app_name='supervisor', api_name='complete_task'))
```

[[[ user ]]]
Output:
```
INFO apis.api_docs.show_api_doc {"app_name": "supervisor", "api_name": "complete_task"} -> {"app_name": "supervisor", "api_name": "complete_task", "path": "/supervisor/complete_task", "method": "POST", "description": "Signal that the task is finished. Call once, last, with the answer if the task asks a question.", "parameters": [{"name": "answer", "required": false, "type": "string"}], "response_schemas": {"success": {"message": "string"}, "failure": "an exception is raised"}}
{
 "app_name": "supervisor",
 "api_name": "complete_task",
 "path": "/supervisor/complete_task",
 "method": "POST",
 "description": "Signal that the task is finished. Call once, last, with the answer if the task asks a question.",
 "parameters": [
  {
   "name": "answer",
   "required": false,
   "type": "string"
  }
 ],
 "response_schemas": {
  "success": {
   "message": "string"
  },
  "failure": "an exception is raised"
 }
}
```
\end{lstlisting}

\section{Reproducing \accord{}}
\label{app:reproduce-accord}
\label{app:accord}

\accord{}~\citep{accord} verifies an agent's action before it runs. A second
model, the grounding agent, receives the task, the agent's trajectory and the
code the agent is about to execute; it may issue read-only calls to observe the
environment, and it returns \code{APPROVE} or \code{REJECT} together with a note
of what it observed. On \code{REJECT} the code is discarded, the note is
appended to the agent's history, and the agent drafts again. The paper releases
no code, so we re-implemented the method from its description, its Algorithm~1
and its prompt appendix. This section records what the paper fixes, what it
leaves open and the choice we made in each case, the prompt and model
configuration, and how our numbers compare with the reported ones.

\paragraph{What the paper specifies}
The backbone and its decoding (Qwen3.5-27B in FP8, temperature 0); the
grounding-agent prompt (its Appendix A.5, reproduced verbatim in
\S\ref{app:prompts-accord}); the inputs to that prompt, namely the pending code,
the context and the task; three policy items added to the agent's instructions,
numbered (10) to (12) in the paper: inspect raw content before writing parsing
logic, re-examine the API specification when a response is not as expected, and
list the APIs of every app before starting the task; and the control flow of
Algorithm~1: an action that calls a sensitive API is withheld, the grounding
agent observes and decides, and on \code{REJECT} the attempt is removed from the
agent's history and the grounding agent's evidence is appended in its place.

\paragraph{What it leaves open, and what we chose}
\Cref{tab:accord-choices} lists the remaining decisions and our choice for each.

\begin{table}[H]
\centering
\small
\providecommand{\rcell}{\raggedright\let\\\tabularnewline}
\caption{Decisions the paper leaves open and the choice taken in our
implementation of \accord{}.}
\label{tab:accord-choices}
\begin{tabular}{@{}p{0.27\textwidth}p{0.68\textwidth}@{}}
\toprule
Left open by the paper & Our choice \\
\midrule
Which calls are sensitive &
\rcell Every endpoint whose HTTP method is not \code{GET} (\code{POST}, \code{PATCH}, \code{DELETE}), plus \code{supervisor.complete\_task}. The set is computed from the code before execution, by a syntax-tree walk over \code{apis.<app>.<endpoint>(} calls. \\
Grounding-agent model &
\rcell The same model and decoding as the agent. \\
Grounding-agent memory &
\rcell One chat per task, kept across verifications. Each verification after the first shows only the steps taken since the previous one. \\
Where observations run &
\rcell In the agent's own namespace, so the grounding agent can read the agent's variables. Write endpoints raise instead of executing, and the observation code runs inside a function so that it cannot rebind the agent's variables. \\
Observation budget &
\rcell Five observation rounds per verification. A reply with neither code nor a decision is answered with a one-line reminder of the output format and costs a round. When the budget is exhausted the decision is forced to \code{APPROVE}. \\
Rejections per task &
\rcell Three. After the third, a \code{REJECT} is ignored and the code executes. \\
Message on \code{REJECT} &
\rcell The \code{OBSERVE} note, appended to the last user turn as ``Your previous code was NOT executed. Before executing sensitive API calls, the environment was observed and the following was reported: \ldots\ Revise your code accordingly.'' (\S\ref{app:prompts-accord}). The discarded attempt is removed, as Algorithm~1 specifies; the wording around the note is ours. \\
Layout of the grounding input &
\rcell Four labeled blocks: the task instruction, the agent trajectory, the code about to execute, and the triggered sensitive APIs with their HTTP methods (\S\ref{app:prompts-accord}). \\
Policy item (12) &
\rcell The item tells the agent to begin by calling \code{show\_api\_descriptions(app\_name="")}, which lists the APIs of every app. Left to the model, it repeats this call until the context is full, so we issue the call for it as the first step of every task and let the environment answer the empty name with the list of all apps. \\
Step limit &
\rcell None. An episode ends when the agent calls \code{complete\_task} or when the context limit is reached. \\
\bottomrule
\end{tabular}
\end{table}

\paragraph{Base prompt and demonstrations}
The paper names no base prompt for the main agent, and its policy items are numbered from (10). We use the ACE
generator prompt~\citep{ace} with its seed playbook, whose key instructions end at (8), and insert the three policy
items after the last key instruction. AppWorld's official prompt, the other public candidate, ends its demonstration
by passing an answer to \code{complete\_task}, and with it the agent passes answers on tasks that expect none. The
ACE prompt carries three demonstrations. For \Cref{tab:onedemo}, where every method receives one demonstration, we
keep only the first: the other two demonstration blocks are removed, and the playbook, the key instructions, and
the task turn are unchanged. \Cref{tab:accord-repro} reports both versions, and \Cref{fig:shots-nothink} varies the
number of demonstrations for every method.

\paragraph{Backbone and decoding}
Qwen3.5-27B in FP8 at temperature 0, for both the agent and the grounding agent.
We report \accord{} with thinking on and with thinking off, since the paper
marks only its 9B model as non-thinking. No cap is placed on generated tokens
other than the 262k-token context. As for every other agent in this paper,
observations are truncated to 20k characters and the history to 400k
characters.

\paragraph{Evaluation}
AppWorld test-normal (168 tasks) and test-challenge (417 tasks), scored by the
official evaluator: TGC is the fraction of tasks whose checks all pass, and SGC
the fraction of scenarios whose tasks all pass. Each row is the mean and sample
standard deviation over three runs.

\paragraph{Result}
\Cref{tab:accord-repro} sets our runs against the paper's Table~1. With the three-demonstration prompt, our
reproduction meets the reported test-challenge numbers with thinking on (71.9 against 69.8 TGC; 59.0 against 53.2
SGC) and falls short on test-normal by 4.3 TGC points (77.2 against 81.5), more than the run-to-run spread. The
budgets, the memory of the grounding agent, and the base prompt are all unstated in the paper, so we cannot
attribute the gap to any one choice in \Cref{tab:accord-choices}. With one demonstration, \accord{} completes more
test-normal tasks (79.4 TGC) and fewer test-challenge tasks (63.5 TGC) than with three. The budgets rarely bind: in
the first one-demonstration test-challenge run with thinking, the grounding agent was invoked 2{,}916 times over the
417 tasks and rejected 171 actions; the observation budget ran out 9 times, and the rejection cap overrode 3
rejections.

\begin{table}[H]
\centering
\small
\setlength{\tabcolsep}{5pt}
\caption{\accord{} as reported and as reproduced here (Qwen3.5-27B, FP8, temperature 0), with the ACE prompt's three
demonstrations and with its first demonstration only. The one-demonstration rows are those of \Cref{tab:onedemo}.
The paper does not state how many demonstrations its prompt carried.}
\label{tab:accord-repro}
\begin{tabular}{@{}lllcccc@{}}
\toprule
 & & & \multicolumn{2}{c}{test-normal} & \multicolumn{2}{c}{test-challenge} \\
\cmidrule(lr){4-5}\cmidrule(lr){6-7}
Thinking & Source & Demos & TGC & SGC & TGC & SGC \\
\midrule
on  & reported~\citep{accord} & -- & 81.5 & 67.9 & 69.8 & 53.2 \\
on  & reproduced (3 runs) & 3 & 77.2{\scriptsize$\,\pm$1.4} & 62.5{\scriptsize$\,\pm$4.7} & 71.9{\scriptsize$\,\pm$1.0} & 59.0{\scriptsize$\,\pm$1.9} \\
on  & reproduced (3 runs) & 1 & 79.4{\scriptsize$\,\pm$1.8} & 65.5{\scriptsize$\,\pm$5.7} & 63.5{\scriptsize$\,\pm$1.7} & 48.4{\scriptsize$\,\pm$1.5} \\
off & reproduced (3 runs) & 3 & 76.4{\scriptsize$\,\pm$3.9} & 62.5{\scriptsize$\,\pm$3.1} & 69.3{\scriptsize$\,\pm$1.8} & 52.3{\scriptsize$\,\pm$1.5} \\
off & reproduced (3 runs) & 1 & 76.2{\scriptsize$\,\pm$1.2} & 59.5{\scriptsize$\,\pm$2.7} & 55.0{\scriptsize$\,\pm$0.4} & 44.8{\scriptsize$\,\pm$2.3} \\
\bottomrule
\end{tabular}
\end{table}

\section{\accord{} prompts}
\label{app:prompts-accord}

\lstdefinestyle{prompt}{
  basicstyle=\ttfamily\scriptsize,
  columns=fullflexible,
  keepspaces=true,
  breaklines=true,
  breakatwhitespace=true,
  frame=single,
  framerule=0.4pt,
  xleftmargin=4pt,
  xrightmargin=4pt,
  aboveskip=6pt,
  belowskip=6pt,
  showstringspaces=false,
  upquote=true,
  inputencoding=utf8,
  extendedchars=true,
  literate={—}{{---}}1 {–}{{--}}1,
  captionpos=b,
}
\renewcommand{\lstlistingname}{Prompt}

The \accord{} agent is the ACE generator prompt~\citep{ace} (with its three demonstrations, or its first one
only in \Cref{tab:onedemo}; \Cref{app:reproduce-accord}), with the
paper's policy items (10) to (12) inserted after its eighth key instruction
(Prompt~\ref{lst:accord-policy}) and the seed playbook of
Prompt~\ref{lst:accord-playbook} in its playbook slot. The grounding agent runs
on Prompt~\ref{lst:accord-grounding}, the paper's Appendix A.5 copied verbatim
with its three worked examples; its final turn is filled with the reflection
context of Prompt~\ref{lst:accord-context}, whose layout is ours. When the
decision is \code{REJECT}, the agent receives the \code{OBSERVE} note wrapped as
in Prompt~\ref{lst:accord-reject}.

\begin{lstlisting}[style=prompt, label={lst:accord-grounding}, caption={The grounding-agent prompt, Appendix A.5 of the \accord{} paper, verbatim. The placeholder on the last line is replaced by the reflection context of Prompt~\ref{lst:accord-context}.}]
USER:
You are a verification agent. Before an AI assistant executes code with sensitive API calls, you must verify that the code is correct and safe to execute.

## Core principle: Observe, then verify, then decide

Your primary tool is **observation** — executing read-only Python code to inspect the actual environment state. Do NOT reason about correctness from the code alone. Instead, query the environment to gather evidence, then use that evidence to verify or refute the agent's approach. Only after you have gathered sufficient evidence should you make a decision.

## How to observe

Write read-only Python code in ```python ...``` blocks. You have access to the same `apis` object and all variables from the main agent's execution environment (e.g., access_token, results from prior steps).

**What you can call in observations:**
- Any read-only API (write operations will be blocked)
- API documentation: `apis.api_docs.show_api_descriptions(app_name=...)` and `apis.api_docs.show_api_doc(app_name=..., api_name=...)` to check API specifications

**Leverage the trajectory before observing:**
You have access to the agent's full execution trajectory, including all intermediate print outputs and API responses. Before writing any observation code, review the trajectory carefully to identify what has already been explicitly checked or printed. If the trajectory already contains the evidence you need (e.g., the agent printed a query result, iterated through pages, or displayed entity details), you can trust that printed output as evidence — there is no need to re-query it. Only observe what is **missing or ambiguous** in the trajectory.

**Observation strategies (apply only to gaps not already covered by the trajectory):**
- **Re-query the data source**: If the agent operated on data it never actually inspected in the trajectory, re-fetch the raw data yourself to verify that the agent's logic matches the real data structure and content.
- **Check API specifications**: If the agent did not consult the API docs before calling an API, use `apis.api_docs.show_api_doc()` to verify it used correct parameter names, types, and values.
- **Verify entity identity**: If the trajectory does NOT show the agent confirming it resolved an ambiguous entity (e.g., which "John" or which "note"), observe the environment to verify.
- **Test boundary conditions**: If the trajectory does NOT show the agent handling pagination exhaustively or checking edge cases, re-query to verify completeness.

## Verification checklist

For each verification item below, use observation to gather evidence before making a judgment.

### 1. Constraint satisfaction
Re-read the task instruction word by word. Extract and QUOTE every explicit constraint verbatim (e.g., recipients, amounts, dates, conditions, filters, set operations like "X that are in Y"). Do NOT paraphrase — copy the exact wording. Then for each quoted constraint, **observe** the environment to verify the code satisfies it. Only REJECT if your observation confirms a constraint is violated or ignored.

### 2. Ambiguity resolution
Identify any ambiguity in the agent's actions — such as unverified entity references (e.g., which "John"? which "meeting"?), assumed IDs, or implicit choices made without evidence. Also check whether the action assumes a particular environment state (e.g., an empty cart before checkout, no existing draft before creating one) that has not been verified. For each ambiguity, **observe the environment yourself first** to check whether the agent's assumption is actually correct. Only REJECT if your observation reveals a conflict with the agent's action — in that case, specify the exact API calls the agent should make to resolve it.

## Workflow

1. Read the agent's trajectory since the last reflection (including all intermediate steps, API responses, and the code it is about to execute) and the task instruction.
2. Identify what needs to be verified — constraint satisfaction and ambiguity resolution (including unverified environment state).
3. **Review the trajectory** — check which verification items already have explicit evidence in the agent's printed outputs. Note these as "covered by trajectory".
4. **Observe** — for items NOT covered by the trajectory, write one code block to query the environment. Wait for the result.
5. Repeat step 4 as needed until you have enough evidence for all items.
6. **Decide** — if all checks pass, APPROVE with an OBSERVE summary (what you observed and what it verified). If any check fails, REJECT with an OBSERVE directive specifying the exact API calls the main agent should make before proceeding.

## Output format

When observing (one code block per message):
```python
# Describe what you are checking
print(...)
```

When deciding (only after sufficient observation):
DECISION: APPROVE
OBSERVE: <what you observed (which APIs you called, what data you checked) and what it verified (which constraints and assumptions were confirmed)>

or

DECISION: REJECT
OBSERVE: <specify the API calls the main agent should make and what information to pay attention to — do NOT include your own judgment or conclusion>

## Key rules

- NEVER output observation code and a DECISION in the same message. Observe first, wait for results, then decide in a separate message.
- Write one code block per observation step. Do not combine multiple unrelated checks.
- You may take multiple observation steps before deciding.
- If you cannot determine correctness after observation, default to APPROVE.
- Do NOT attempt to fix or rewrite the code. Only APPROVE or REJECT.

## Examples

----------------------------------------------

USER:
[Reflection context omitted: contains task instruction, full agent trajectory, code about to execute, and triggered sensitive APIs]

ASSISTANT:
The agent is about to execute:
```python
message = "Here are my David Fincher movie recommendations: Fight Club, The Social Network"
apis.phone.send_text_message(access_token=phone_access_token, phone_number=christopher_phone, message=message)
```

Task: "Christopher has asked for my movie recommendations via phone text message. Reply to them with a list of comma-separated movie titles from my Simple Note account as per their request."

Constraints to verify:
- "Christopher has asked ... via phone text message" — need to observe the original message to confirm what Christopher asked for
- "movie titles from my Simple Note account" — need to observe the note content to confirm the movies match
- "as per their request" — need to confirm the movies satisfy Christopher's specific request

Let me observe Christopher's original message first.
```python
messages = apis.phone.search_text_messages(access_token=phone_access_token, phone_number=christopher_phone, query="movie")
print(messages)
```

USER:
Output:
```
[
  {
    "text_message_id": 16794,
    "sender": {
      "name": "Christopher Burch",
      "phone_number": "4275730168"
    },
    "message": "Hey, I'm planning to watch a movie. Do you have any recommendations for a movie from David Fincher?",
    "sent_at": "2023-05-18T10:17:40"
  }
]
```

ASSISTANT:
Christopher specifically asked for "recommendations for a movie from David Fincher". The agent's message includes "Fight Club" and "The Social Network". Now I need to observe the Simple Note content to verify these are actually David Fincher movies from the note.
```python
movie_note = apis.simple_note.show_note(access_token=simple_note_access_token, note_id=1608)
print(movie_note['content'])
```

USER:
Output:
```
# Movie Recommendations

Interstellar
- director: Christopher Nolan
- genre: Sci-Fi, Drama

Fight Club
- director: David Fincher
- genre: Drama

The Social Network
- director: David Fincher
- genre: Biography, Drama

The Godfather
- director: Francis Ford Coppola
- genre: Crime, Drama
```

ASSISTANT:
DECISION: APPROVE
OBSERVE: Called `phone.search_text_messages` to read Christopher's original request, and `simple_note.show_note` to read the note content. Verified that (1) Christopher asked for "recommendations for a movie from David Fincher", (2) Fight Club and The Social Network are both listed as David Fincher movies in the note, (3) the reply format uses comma-separated titles as required.

----------------------------------------------

USER:
[Reflection context omitted: contains task instruction, full agent trajectory, code about to execute, and triggered sensitive APIs]

ASSISTANT:
The agent is about to execute:
```python
apis.supervisor.complete_task(answer=23)
```

Task: "How many playlists do I have in Spotify?"

I need to independently verify the count of 23. Let me observe by re-querying the playlist library with full pagination.
```python
playlists = []
page_index = 0
while True:
    page = apis.spotify.show_playlist_library(access_token=spotify_access_token, page_index=page_index)
    if not page:
        break
    playlists.extend(page)
    page_index += 1
print(f"Total playlists: {len(playlists)}, pages iterated: {page_index}")
```

USER:
Output:
```
Total playlists: 23, pages iterated: 3
```

ASSISTANT:
DECISION: APPROVE
OBSERVE: Called `spotify.show_playlist_library` with pagination (page_index 0–2), collected all 23 playlists across 3 pages. Verified that the total count matches the agent's answer of 23.

----------------------------------------------

USER:
[Reflection context omitted: contains task instruction, full agent trajectory, code about to execute, and triggered sensitive APIs]

ASSISTANT:
The agent is about to execute:
```python
apis.venmo.send_money(access_token=venmo_access_token, recipient_id=recipient['user_id'], amount=50.00, note="Dinner split")
```

Task: "Split the $120 dinner bill equally among me and my 3 friends on Venmo."

Constraints to verify:
- "Split the $120 dinner bill equally" — need to verify the amount is correct
- "among me and my 3 friends" — 4 people total, each share = $30, but agent sends $50

The amount looks wrong. Let me observe to confirm.
```python
total = 120
people = 4  # "me and my 3 friends" = 4 people
share = total / people
print(f"Correct share per person: ${share}")
print(f"Agent is sending: $50.00")
```

USER:
Output:
```
Correct share per person: $30.0
Agent is sending: $50.00
```

ASSISTANT:
DECISION: REJECT
OBSERVE: Before sending money, call `apis.venmo.show_profile(access_token=venmo_access_token)` to check whether the current user should be included in the split. Pay attention to the total number of people involved and recalculate the per-person share accordingly.

----------------------------------------------

USER:
Now verify the following code that the agent is about to execute.

{{ reflect_context }}
\end{lstlisting}

\begin{lstlisting}[style=prompt, label={lst:accord-context}, caption={The reflection context rendered into the grounding agent's final turn. On the first verification of a task the heading reads ``Agent trajectory:'' and the whole trajectory follows; later verifications show only the steps since the previous one.}]
Task instruction: {task}

Agent trajectory (new steps since the last reflection):
{steps}
Code about to execute:
```python
{code}
```

Triggered sensitive APIs: apis.{app}.{endpoint} ({METHOD}), ...
\end{lstlisting}

\begin{lstlisting}[style=prompt, label={lst:accord-reject}, caption={The message appended to the agent's last user turn when the grounding agent rejects; the discarded attempt itself is removed from the history.}]
Your previous code was NOT executed. Before executing sensitive API calls, the environment was observed and the following was reported:
{observe}

Revise your code accordingly.
\end{lstlisting}

\begin{lstlisting}[style=prompt, label={lst:accord-policy}, caption={Policy items (10) to (12) of the \accord{} paper, inserted after key instruction (8) of the agent prompt.}]
(10) If the content or format of returned information is uncertain, inspect the actual raw content first (e.g., via read/show APIs) and do not guess structure or write parsing logic based on assumptions.

(11) If the content of returned information is not as expected, re-examine the API specification, your request parameters, and any assumptions you made about the response structure. Adjust your approach accordingly and verify the result is correct before proceeding to the next step.

(12) Before executing the task, you MUST call show_api_descriptions(app_name="") to get the complete API list for ALL apps at once. Do NOT selectively check only apps you think are relevant—you must examine every app's APIs exhaustively.
\end{lstlisting}

\begin{lstlisting}[style=prompt, label={lst:accord-playbook}, caption={The seed playbook placed in the playbook slot of the ACE generator prompt, used by \accord{} and by \react{}; our agent uses no playbook.}]
## STRATEGIES AND HARD RULES
[shr-00001] Make sure to end code blocks with ``` followed by a newline(\n).
[shr-00005] Always look at API specifications (using apis.api_docs.show_api_doc) before calling an API.
[shr-00006] Write small chunks of code and only one chunk of code in every step. Make sure everything is working correctly before making any irreversible change.

## APIs TO USE FOR SPECIFIC INFORMATION
[api-00004] You can use the "supervisor" app to get information about my accounts and use the "phone" app to get information about friends and family.

## USEFUL CODE SNIPPETS AND TEMPLATES

## COMMON MISTAKES AND CORRECT STRATEGIES

## PROBLEM-SOLVING HEURISTICS AND WORKFLOWS
[psw-00002] Remember you can use the variables in your code in subsequent code blocks.
[psw-00007] Many APIs return items in "pages". Make sure to run through all the pages by looping over `page_index`.

## VERIFICATION CHECKLIST

## TROUBLESHOOTING AND PITFALLS:

## OTHERS
[misc-00003] Remember that the email addresses, access tokens and variables (e.g. spotify_password) in the example above are not valid anymore.
[misc-00008] Once you have completed the task, make sure to call apis.supervisor.complete_task(). If the task asked for some information, return it as the answer argument, i.e. call apis.supervisor.complete_task(answer=<answer>). Many tasks do not require an answer, so in those cases, just call apis.supervisor.complete_task() i.e. do not pass any argument.
\end{lstlisting}

\raggedbottom
\section{Cost measurement}
\label{app:cost}

\hcy{\paragraph{Token counts}
For every model call, the serving engine reports the number of input tokens it received and the number of tokens it
generated: SGLang for the Qwen backbones, which we serve ourselves, and the API provider (OpenRouter) for GLM-5.3-flash
and MiniMax~M2.7. We sum both counts over all model calls of an episode, including the verification calls of our
agent and the grounding calls of \accord{}. Input tokens count every token sent on every call, including a prefix
that a serving cache could reuse; generated tokens include thinking tokens. For each task we average over the runs
of a configuration, and then average over the tasks of the split. When a task has more than one trial in a run
(a re-run after a lost worker), we use the latest.

\paragraph{Cost measures}
The first measure is the number of generated tokens per task. The second prices input and generated tokens per
task at one common price, the OpenRouter list price of Qwen3.5-27B (\$0.195 per million input tokens and \$1.56 per
million generated tokens, September 2026), so that the measure reflects how many tokens a harness uses rather than
which model is cheap. \Cref{fig:cost-harness} divides both measures, and the task goal completion, by those of
\react{} with thinking on the same backbone. An agent lies on the frontier of its backbone if no other agent on the
same backbone costs no more and completes more tasks.

\paragraph{All measured points}
\Cref{tab:cost-harness-full} lists every point on both test splits, including Qwen3-30B-A3B, which
\Cref{fig:cost-harness} omits. On Qwen3.5-27B, the runs are those of \Cref{tab:onedemo} (one demonstration, three
runs). The other backbones have one run each, and on the two API backbones the baselines use the released
three-demonstration prompt, so comparisons across backbones mix configurations; comparisons within a backbone do
not.}

\begin{table}[H]
\centering\small\setlength{\tabcolsep}{3pt}
\caption{Task goal completion and cost of every backbone and harness on AppWorld, with the backbone model's thinking
on~(\cmark) and off~(\xmark). Tokens: generated tokens per task, in thousands. Cost: input and generated tokens per
task priced at the common price, in dollars. In parentheses, the ratio to \react{} with thinking on the same
backbone. $\uparrow$: higher is better; $\downarrow$: lower is better. Mean $\pm$ standard deviation of TGC over
three runs on Qwen3.5-27B; one run on the other backbones. Bold is the best TGC within its thinking block,
underline the runner-up.}
\label{tab:cost-harness-full}
\begin{tabular*}{\linewidth}{@{\extracolsep{\fill}}clllllll@{}}
\toprule
 & & \multicolumn{3}{l}{Test (normal)} & \multicolumn{3}{l}{Test (challenge)} \\
\cmidrule(l{1pt}r{1pt}){3-5}\cmidrule(l{1pt}r{1pt}){6-8}
 & Method & TGC$\uparrow$ & Tokens$\downarrow$ & Cost$\downarrow$ & TGC$\uparrow$ & Tokens$\downarrow$ & Cost$\downarrow$ \\
\midrule
\multicolumn{8}{@{}l}{\itshape Qwen3.5-27B} \\
\multirow{3}{*}{\cmark{}} & \react{} & 72.4{\scriptsize$\,\pm$1.9} & \phantom{00}2.7{\scriptsize$\,$(1.0$\times$)} & 0.032{\scriptsize$\,$(1.0$\times$)} & 58.4{\scriptsize$\,\pm$0.8} & \phantom{00}3.6{\scriptsize$\,$(1.0$\times$)} & 0.047{\scriptsize$\,$(1.0$\times$)} \\
 & \accord{} & \underline{79.4}{\scriptsize$\,\pm$1.8} & \phantom{0}12.9{\scriptsize$\,$(4.7$\times$)} & 0.092{\scriptsize$\,$(2.9$\times$)} & \underline{63.5}{\scriptsize$\,\pm$1.7} & \phantom{0}10.5{\scriptsize$\,$(2.9$\times$)} & 0.130{\scriptsize$\,$(2.7$\times$)} \\
\cmidrule{2-8}
 & \ours{} & \textbf{88.3}{\scriptsize$\,\pm$3.0} & \phantom{0}17.7{\scriptsize$\,$(6.5$\times$)} & 0.235{\scriptsize$\,$(7.4$\times$)} & \textbf{84.7}{\scriptsize$\,\pm$1.3} & \phantom{0}22.9{\scriptsize$\,$(6.4$\times$)} & 0.447{\scriptsize$\,$(9.4$\times$)} \\
\cmidrule{1-8}
\multirow{3}{*}{\xmark{}} & \react{} & 70.0{\scriptsize$\,\pm$2.5} & \phantom{00}2.4{\scriptsize$\,$(0.9$\times$)} & 0.028{\scriptsize$\,$(0.9$\times$)} & 48.0{\scriptsize$\,\pm$0.8} & \phantom{00}2.2{\scriptsize$\,$(0.6$\times$)} & 0.044{\scriptsize$\,$(0.9$\times$)} \\
 & \accord{} & \underline{76.2}{\scriptsize$\,\pm$1.2} & \phantom{00}3.9{\scriptsize$\,$(1.4$\times$)} & 0.079{\scriptsize$\,$(2.5$\times$)} & \underline{55.0}{\scriptsize$\,\pm$0.4} & \phantom{00}4.9{\scriptsize$\,$(1.4$\times$)} & 0.112{\scriptsize$\,$(2.4$\times$)} \\
\cmidrule{2-8}
 & \ours{} & \textbf{84.1}{\scriptsize$\,\pm$1.5} & \phantom{00}5.3{\scriptsize$\,$(2.0$\times$)} & 0.195{\scriptsize$\,$(6.1$\times$)} & \textbf{81.5}{\scriptsize$\,\pm$2.9} & \phantom{00}7.3{\scriptsize$\,$(2.1$\times$)} & 0.366{\scriptsize$\,$(7.7$\times$)} \\
\midrule
\multicolumn{8}{@{}l}{\itshape Qwen3-30B-A3B} \\
\multirow{3}{*}{\cmark{}} & \react{} & \underline{32.1}\phantom{{\scriptsize$\,\pm$0.0}} & \phantom{0}22.4{\scriptsize$\,$(1.0$\times$)} & 0.055{\scriptsize$\,$(1.0$\times$)} & \underline{15.8}\phantom{{\scriptsize$\,\pm$0.0}} & \phantom{0}29.9{\scriptsize$\,$(1.0$\times$)} & 0.092{\scriptsize$\,$(1.0$\times$)} \\
 & \accord{} & 29.8\phantom{{\scriptsize$\,\pm$0.0}} & \phantom{0}30.0{\scriptsize$\,$(1.3$\times$)} & 0.101{\scriptsize$\,$(1.8$\times$)} & 13.7\phantom{{\scriptsize$\,\pm$0.0}} & \phantom{0}47.9{\scriptsize$\,$(1.6$\times$)} & 0.175{\scriptsize$\,$(1.9$\times$)} \\
\cmidrule{2-8}
 & \ours{} & \textbf{52.4}\phantom{{\scriptsize$\,\pm$0.0}} & \phantom{0}43.8{\scriptsize$\,$(2.0$\times$)} & 0.193{\scriptsize$\,$(3.5$\times$)} & \textbf{34.8}\phantom{{\scriptsize$\,\pm$0.0}} & \phantom{0}54.0{\scriptsize$\,$(1.8$\times$)} & 0.292{\scriptsize$\,$(3.2$\times$)} \\
\midrule
\multicolumn{8}{@{}l}{\itshape Qwen3-4B} \\
\multirow{3}{*}{\cmark{}} & \react{} & \underline{23.2}\phantom{{\scriptsize$\,\pm$0.0}} & \phantom{0}60.1{\scriptsize$\,$(1.0$\times$)} & 0.134{\scriptsize$\,$(1.0$\times$)} & \underline{12.2}\phantom{{\scriptsize$\,\pm$0.0}} & \phantom{0}64.6{\scriptsize$\,$(1.0$\times$)} & 0.154{\scriptsize$\,$(1.0$\times$)} \\
 & \accord{} & 14.9\phantom{{\scriptsize$\,\pm$0.0}} & 108.6{\scriptsize$\,$(1.8$\times$)} & 0.287{\scriptsize$\,$(2.1$\times$)} & \phantom{0}8.6\phantom{{\scriptsize$\,\pm$0.0}} & 109.4{\scriptsize$\,$(1.7$\times$)} & 0.269{\scriptsize$\,$(1.8$\times$)} \\
\cmidrule{2-8}
 & \ours{} & \textbf{41.7}\phantom{{\scriptsize$\,\pm$0.0}} & \phantom{0}77.7{\scriptsize$\,$(1.3$\times$)} & 0.293{\scriptsize$\,$(2.2$\times$)} & \textbf{27.6}\phantom{{\scriptsize$\,\pm$0.0}} & \phantom{0}96.0{\scriptsize$\,$(1.5$\times$)} & 0.359{\scriptsize$\,$(2.3$\times$)} \\
\midrule
\multicolumn{8}{@{}l}{\itshape GLM-5.3-Flash} \\
\multirow{3}{*}{\cmark{}} & \react{} & \underline{83.9}\phantom{{\scriptsize$\,\pm$0.0}} & \phantom{00}1.3{\scriptsize$\,$(1.0$\times$)} & 0.027{\scriptsize$\,$(1.0$\times$)} & \underline{77.5}\phantom{{\scriptsize$\,\pm$0.0}} & \phantom{00}1.6{\scriptsize$\,$(1.0$\times$)} & 0.041{\scriptsize$\,$(1.0$\times$)} \\
 & \accord{} & 82.1\phantom{{\scriptsize$\,\pm$0.0}} & \phantom{00}2.6{\scriptsize$\,$(2.0$\times$)} & 0.058{\scriptsize$\,$(2.1$\times$)} & 76.7\phantom{{\scriptsize$\,\pm$0.0}} & \phantom{00}3.3{\scriptsize$\,$(2.0$\times$)} & 0.076{\scriptsize$\,$(1.9$\times$)} \\
\cmidrule{2-8}
 & \ours{} & \textbf{91.1}\phantom{{\scriptsize$\,\pm$0.0}} & \phantom{00}6.9{\scriptsize$\,$(5.3$\times$)} & 0.226{\scriptsize$\,$(8.3$\times$)} & \textbf{83.5}\phantom{{\scriptsize$\,\pm$0.0}} & \phantom{00}8.6{\scriptsize$\,$(5.2$\times$)} & 0.363{\scriptsize$\,$(8.9$\times$)} \\
\midrule
\multicolumn{8}{@{}l}{\itshape MiniMax M2.7} \\
\multirow{3}{*}{\cmark{}} & \react{} & 61.3\phantom{{\scriptsize$\,\pm$0.0}} & \phantom{00}3.1{\scriptsize$\,$(1.0$\times$)} & 0.038{\scriptsize$\,$(1.0$\times$)} & 53.0\phantom{{\scriptsize$\,\pm$0.0}} & \phantom{00}3.5{\scriptsize$\,$(1.0$\times$)} & 0.053{\scriptsize$\,$(1.0$\times$)} \\
 & \accord{} & \underline{71.4}\phantom{{\scriptsize$\,\pm$0.0}} & \phantom{00}6.9{\scriptsize$\,$(2.2$\times$)} & 0.090{\scriptsize$\,$(2.4$\times$)} & \underline{63.1}\phantom{{\scriptsize$\,\pm$0.0}} & \phantom{00}8.1{\scriptsize$\,$(2.3$\times$)} & 0.121{\scriptsize$\,$(2.3$\times$)} \\
\cmidrule{2-8}
 & \ours{} & \textbf{77.4}\phantom{{\scriptsize$\,\pm$0.0}} & \phantom{0}12.8{\scriptsize$\,$(4.1$\times$)} & 0.199{\scriptsize$\,$(5.2$\times$)} & \textbf{73.9}\phantom{{\scriptsize$\,\pm$0.0}} & \phantom{0}14.4{\scriptsize$\,$(4.1$\times$)} & 0.302{\scriptsize$\,$(5.8$\times$)} \\
\bottomrule
\end{tabular*}

\end{table}

\section{Qualitative examples}
\label{app:qualitative}

\hcy{\Cref{tab:qual-alarm,tab:qual-wrench} show two tasks from the AppWorld test-challenge split, run with
Qwen3.5-27B and thinking off: \accord{} with the ACE prompt's three demonstrations on the left and \ours{} on the
right, one run each. On both tasks the two agents take the same first steps and propose the same action.
\accord{}'s grounding agent reads the environment to confirm the action and approves it once what it reads agrees
with the action. \ours{} instead asks what would make the action wrong and reads for that. In the first task it
finds a third cancellation, worded without the search terms; in the second it finds that a returned wrench has the
product type ``crescent wrench'' although its name does not say so. \ours{} then revises the action and completes
both tasks, while \accord{} fails both. Model outputs are verbatim, shortened with [...]; green marks a right
verification, red a wrong one, and blue the ground truth.}

\begin{table}[H]
\centering\footnotesize\setlength{\tabcolsep}{5pt}
\caption{Task c1091c7\_1: disable the alarms of meetings that coworkers canceled by email.}
\label{tab:qual-alarm}
\providecommand{\qwrong}{\textcolor{red}{\ensuremath{\times}}}
\providecommand{\qright}{\textcolor{green!50!black}{\ensuremath{\surd}}}
\providecommand{\qok}[1]{\textcolor{green!50!black}{\textbf{#1}}}
\providecommand{\qbad}[1]{\textcolor{red}{\textbf{#1}}}
\providecommand{\qgt}[1]{\textcolor{blue}{(#1)}}
\providecommand{\qcell}{\raggedright\let\\\tabularnewline}
\providecommand{\qstage}[1]{\par\textbf{#1:} }
\begin{tabular}{@{}p{0.485\linewidth}|p{0.485\linewidth}@{}}
\toprule
\accord{} \qbad{(approves with one alarm still on)} & \ours{} \qok{(refutes and revises)} \\
\midrule
\multicolumn{2}{@{}p{\linewidth}@{}}{\textbf{Task:} I have alarms for all my work meetings. Some of my coworkers have canceled their meeting with me via email. Please disable the corresponding alarms from my phone. \qgt{GT: set \code{enabled=False} on alarms 744, 745, and 743. Three coworkers canceled; Stephen's email is titled ``Skip this time?'' and reads ``Cannot make it.''}} \\
\multicolumn{2}{@{}p{\linewidth}@{}}{\textbf{Shared prefix:} both agents search Gmail for \code{"cancel meeting"}, find the two ``Cancel Meeting?'' threads (Stacy $\to$ alarm 744, Kristin $\to$ alarm 745), and propose \code{phone.delete\_alarm(alarm\_id=744)}.} \\
\midrule
\qcell \textbf{Round 1}
\qstage{Grounding agent} reads the \code{update\_alarm} docs, then: \qok{DECISION: REJECT.} OBSERVE: The task instruction says to ``disable'' the alarms, not delete them. The \code{update\_alarm} API has an \code{enabled} parameter that can be set to \code{false} to disable an alarm without deleting it. [...] \hfill \qright{}
&
\qcell \textbf{Round 1}
\qstage{Suspect} (1) The task asks to `disable' the alarms, but the code calls \code{delete\_alarm} which permanently removes them. (2) The code only deletes the alarm for Stacy (ID 744). [...] alarm labeled `1-1 with Kristin' (ID 745) also needs to be disabled, but this code cell does not address it.
\qstage{Test} reads the \code{update\_alarm} docs $\to$ \qok{both confirmed}. \hfill \qright{} \\
\multicolumn{2}{@{}p{\linewidth}@{}}{\emph{Both revise to} \code{update\_alarm(alarm\_id=744, enabled=False)} and \code{update\_alarm(alarm\_id=745, enabled=False)}.} \\
\midrule
\qcell \textbf{Round 2} --- the two updates run; the agent proposes \code{supervisor.complete\_task()}.
\qstage{Grounding agent} ``Let me verify that no other cancellation emails were missed'' $\to$ probes \code{show\_inbox\_threads(query="cancel")} $\to$ Thread 47823: ``Cancel Meeting?'' from Stacy Maldonado --- handled. Thread 47825: ``Cancel Meeting?'' from Kristin White --- handled. Thread 47831: ``Cancel Reunion'' from Leslie Ball --- this is about a reunion, not a work meeting. Other threads are Amazon notifications, dinner party RSVPs, [...] none are meeting cancellations. \qbad{DECISION: APPROVE. OBSERVE: [...] Confirmed that only two work meeting cancellations were found} (Stacy Maldonado and Kristin White), both of which have been handled [...] The task is complete.
\par\smallskip \qwrong{} \qgt{alarm 743 ``1-1 with Stephen'' stays enabled}
&
\qcell \textbf{Round 2} --- the agent proposes the two \code{update\_alarm} calls.
\qstage{Expect} Disables alarms 744 (`1-1 with Stacy') and 745 (`1-1 with Kristin') by setting \code{enabled=False}, matching the two meeting cancellations found in emails [...]
\qstage{Suspect} [...] \qok{The real risk is if there are other cancellation emails not captured by the query `cancel meeting' (e.g., `reschedule', `postpone', `cannot make it') that correspond to alarms.} If other coworkers canceled via different phrasing, their alarms would remain enabled [...] \code{ref: phone.update\_alarm.alarm\_id}. \emph{[Two further suspicions, on the label--sender mapping and on the meaning of `disable', are dismissed in the test and omitted here.]}
\qstage{Test} probes \code{show\_inbox\_threads(query="reschedule OR postpone OR cannot make OR conflict")} $\to$ thread 47821 ``Skip this time?''; then \code{show\_thread(47821)} $\to$ sender Stephen Mccoy, ``Cannot make it.'', and alarm 743 ``1-1 with Stephen'' is enabled.
\qstage{Verdict} \qok{confirmed} --- Email thread 47821 from Stephen Mccoy says `Cannot make it' (cancellation), and alarm 743 `1-1 with Stephen' exists but was not included in the proposed code.
\qstage{Reframe} disables 744, 745, and 743.
\par\smallskip \qright{} \qgt{all three alarms disabled} \\
\bottomrule
\end{tabular}

\end{table}

\begin{table}[H]
\centering\footnotesize\setlength{\tabcolsep}{5pt}
\caption{Task 3650990\_3: has the user returned a crescent wrench this or last month?}
\label{tab:qual-wrench}
\providecommand{\qwrong}{\textcolor{red}{\ensuremath{\times}}}
\providecommand{\qright}{\textcolor{green!50!black}{\ensuremath{\surd}}}
\providecommand{\qok}[1]{\textcolor{green!50!black}{\textbf{#1}}}
\providecommand{\qbad}[1]{\textcolor{red}{\textbf{#1}}}
\providecommand{\qgt}[1]{\textcolor{blue}{(#1)}}
\providecommand{\qcell}{\raggedright\let\\\tabularnewline}
\providecommand{\qstage}[1]{\par\textbf{#1:} }
\begin{tabular}{@{}p{0.485\linewidth}|p{0.485\linewidth}@{}}
\toprule
\accord{} \qbad{(sees the wrench, approves ``no'')} & \ours{} \qok{(asks what would make ``no'' wrong)} \\
\midrule
\multicolumn{2}{@{}p{\linewidth}@{}}{\textbf{Task:} Have I returned any crescent wrench on amazon in this or last month? Say yes or no. \qgt{GT: yes --- ``Craftsman 8-Inch Adjustable Wrench'', \code{product\_type} ``crescent wrench'', returned on 2023-04-28.}} \\
\multicolumn{2}{@{}p{\linewidth}@{}}{\textbf{Shared prefix:} both agents fetch all 33 returns, keep those from April/May 2023 whose \code{product\_name} contains ``crescent wrench'' (none), and propose \code{supervisor.complete\_task(answer="no")}.} \\
\midrule
\qcell
\qstage{Grounding agent} ``Let me observe to verify the agent's logic by checking if there are any returns with `crescent' or `wrench' in the product name that might have been missed.'' $\to$ probes \code{all\_returns} $\to$ Output: Total wrench returns in April/May 2023: 1 --- Craftsman 8-Inch Adjustable Wrench (initiated: 2023-04-28). Total crescent returns in April/May 2023: 0.
\par\smallskip \qbad{DECISION: APPROVE. OBSERVE:} Verified by re-analyzing all 33 returns that (1) the date range April/May 2023 is correct [...], (2) \qbad{there is 1 wrench return (``Craftsman 8-Inch Adjustable Wrench'') but 0 crescent wrench returns} in the target months, (3) the agent's case-insensitive search for ``crescent wrench'' correctly found no matches. The answer ``no'' is accurate.
\par\smallskip \qwrong{} \qgt{answer: no; GT: yes}
&
\qcell
\qstage{Expect} The code will mark the current task as complete with the answer `no', indicating that the user has not returned any crescent wrench on Amazon in the current or previous month, based on the analysis of all returned items.
\qstage{Suspect} The code searched for the exact phrase `crescent wrench' in the product name. \qok{However, the returned item `Craftsman 8-Inch Adjustable Wrench' is a type of crescent wrench.} [...] The code did not account for synonyms or broader category matching. \code{ref: task}.
\qstage{Test} probes \code{amazon.search\_products(query="crescent wrench")} $\to$ Output: \code{product\_id} 2139, \code{product\_type} ``crescent wrench'', name ``Craftsman 8-Inch Adjustable Wrench'' [...]
\qstage{Verdict} \qok{confirmed} --- Search results show \code{product\_id} 2139 has \code{product\_type} `crescent wrench' but name `Craftsman 8-Inch Adjustable Wrench'. The return list contains this \code{product\_id}.
\qstage{Reframe} re-checks the returns by \code{product\_id} $\to$ found, 2023-04-28 $\to$ \code{complete\_task(answer="yes")}.
\par\smallskip \qright{} \qgt{answer: yes}\\
\bottomrule
\end{tabular}

\end{table}

\end{document}